\documentclass[letterpaper]{article} 
\usepackage{aaai2026}  
\usepackage{times}  
\usepackage{helvet}  
\usepackage{courier}  
\usepackage[hyphens]{url}  
\usepackage{graphicx} 
\usepackage{natbib}  
\usepackage{caption} 
\usepackage{algorithm}

\usepackage{newfloat}
\usepackage{listings}
\DeclareCaptionStyle{ruled}{labelfont=normalfont,labelsep=colon,strut=off} 
\floatstyle{ruled}
\newfloat{listing}{tb}{lst}{}
\floatname{listing}{Listing}
\usepackage{algorithm}
\usepackage{algorithmicx}
\usepackage{algpseudocode}
\usepackage{amsmath, amssymb} 

\usepackage{xcolor}
\usepackage{subcaption}

\usepackage{multirow}
\usepackage{booktabs}
\usepackage{bibentry}

\DeclareMathOperator*{\argmax}{arg\,max}

\title{MosaicDoc: A Large-Scale Bilingual Benchmark for Visually Rich Document Understanding}

\author{
    Ketong Chen,
    Yuhao Chen,
    Yang Xue\thanks{Correspond author}
}
\affiliations{
    School of Electronic and Information Engineering, South China University of Technology\\
    \{eeckt, yxue\}@mail.scut.edu.cn, chardmaplesan@gmail.com
}

\usepackage{bibentry}

\begin{document}

\maketitle

\begin{abstract}
Despite the rapid progress of Vision-Language Models (VLMs), their capabilities are inadequately assessed by existing benchmarks, which are predominantly English-centric, feature simplistic layouts, and support limited tasks. Consequently, they fail to evaluate model performance for Visually Rich Document Understanding (VRDU), a critical challenge involving complex layouts and dense text. To address this, we introduce DocWeaver, a novel multi-agent pipeline that leverages Large Language Models to automatically generate a new benchmark. The result is MosaicDoc, a large-scale, bilingual (Chinese and English) resource designed to push the boundaries of VRDU. Sourced from newspapers and magazines, MosaicDoc features diverse and complex layouts (including multi-column and non-Manhattan), rich stylistic variety from 196 publishers, and comprehensive multi-task annotations (OCR, VQA, reading order, and localization). With 72K images and over 600K QA pairs, MosaicDoc serves as a definitive benchmark for the field. Our extensive evaluation of state-of-the-art models on this benchmark reveals their current limitations in handling real-world document complexity and charts a clear path for future research.
\end{abstract}

\begin{links}
    \link{Code and Datasets}{https://github.com/DOCLAB-SCUT/MosaicDoc}
\end{links}

\section{Introduction}

With the rapid advancement of Document AI, the field is converging on unified, end-to-end models. Document Visual Question Answering (DocVQA) has emerged as a key paradigm, capable of unifying diverse tasks into a single, prompt-based framework~\cite{udop, ye-etal-2023-ureader, feng2023unidoc}. However, the development of these powerful models is fundamentally constrained by the benchmarks used for their evaluation. The central challenge lies in Visually Rich Document Understanding (VRDU), which requires models to comprehend documents with complex layouts, dense text, and diverse visual styles. This remains a domain where current datasets fall critically short.

\begin{figure}[t]
\centering
\includegraphics[width=0.9\columnwidth]{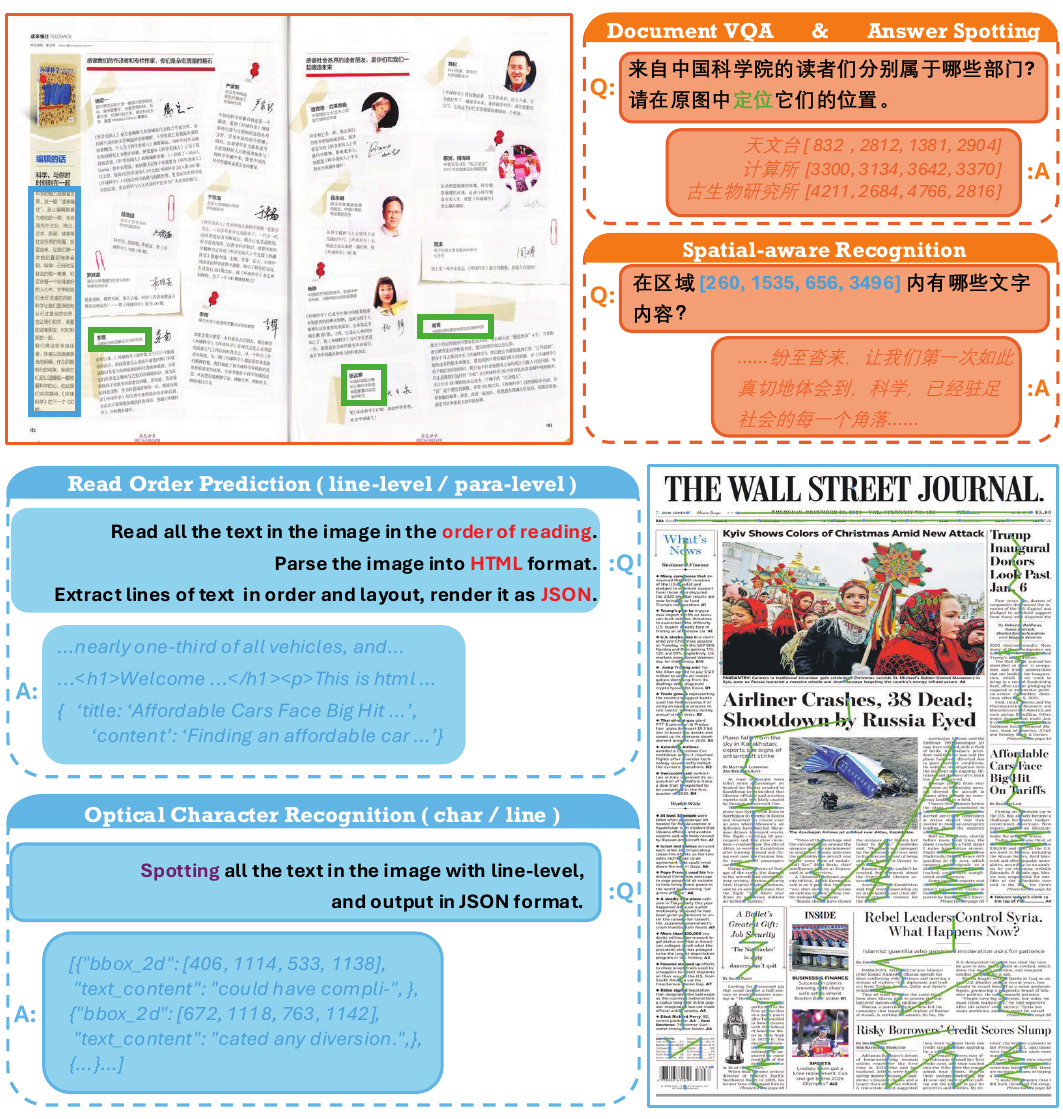}
\caption{Examples of VRDU Tasks in the MosaicDoc Benchmark.}
\label{fig:samples}
\end{figure}

Existing benchmarks, such as VisualMRC~\cite{tanaka2021visualmrc}, and LLM-generated ones like Docmatix~\cite{laurenccon2024docmatix}, are overwhelmingly English-centric and rest upon visually simple documents. They lack stylistic variety and layout complexity representative of real-world artifacts like newspapers and magazines. This limitation is particularly acute for non-English documents. Existing Chinese datasets also fail to capture real-world visual complexity. For instance, XFUND~\cite{xu-etal-2022-xfund} largely replicates the simplistic layouts of its English counterparts, while DuReadervis (Qi et al. 2022), despite using long webpage, utilizes only the visually simple top portions, neglecting their richer content. Crucially, they lack reading order annotations, forcing models to rely on a flawed top-to-bottom reading assumption that fails on multi-column or non-Manhattan layouts. As our experiments confirm, this leads to significant performance degradation in state-of-the-art (SOTA) models.

To address these challenges, we propose DocWeaver, a novel multi-agent pipeline powered by LLMs for generating high-fidelity, multi-task document annotations. DocWeaver employs a modular architecture of specialized agents, where Extractors parse document elements, Generators create diverse questions, and five distinct Hallucination Guardrails rigorously validate each sample for quality and factual consistency. This pipeline explicitly computes and utilizes correct reading order to ensure genuine layout comprehension.
Using this method, we introduce MosaicDoc, a large-scale bilingual benchmark for Visually Rich Document Understanding. Sourced from modern newspapers and magazines, MosaicDoc provides the first comprehensive resource for evaluating models on complex, real-world documents in both Chinese and English. It contains over 600K question-answer pairs across 72K images from 196 publishers, with rich annotations supporting a wide array of tasks, including DocVQA, OCR, reading order prediction, and content-aware localization. We conduct a rigorous evaluation of thirteen SOTA models on MosaicDoc, establishing a new performance baseline.

In summary, our contributions are three-fold:
\begin{itemize}
\item A Novel Automated Pipeline. We introduce \textbf{DocWeaver}, a fully automated multi-agent pipeline that leverages LLMs to generate high-fidelity, multi-task annotations, addressing the criticla data creation bottleneck for visually rich documents.
\item A Challenging New Benchmark. We present \textbf{MosaicDoc}, a large-scale bilingual benchmark from complex newspapers and magazines. It introduces unprecedented diverse in layout and style to facilitate more rigorous research in VRDU.
\item A Rigorous Performance Analysis. We provide a comprehensive benchmark analysis of 13 state-of-the-art models on MosaicDoc. This evaluation reveals systemic weaknesses in current approaches, particularly in handling dense layouts and multi-span reasoning, thereby establishing a new more challenging baseline for the field.
\end{itemize}

\begin{figure*}[t]
\centering
\includegraphics[width=1.0\textwidth]{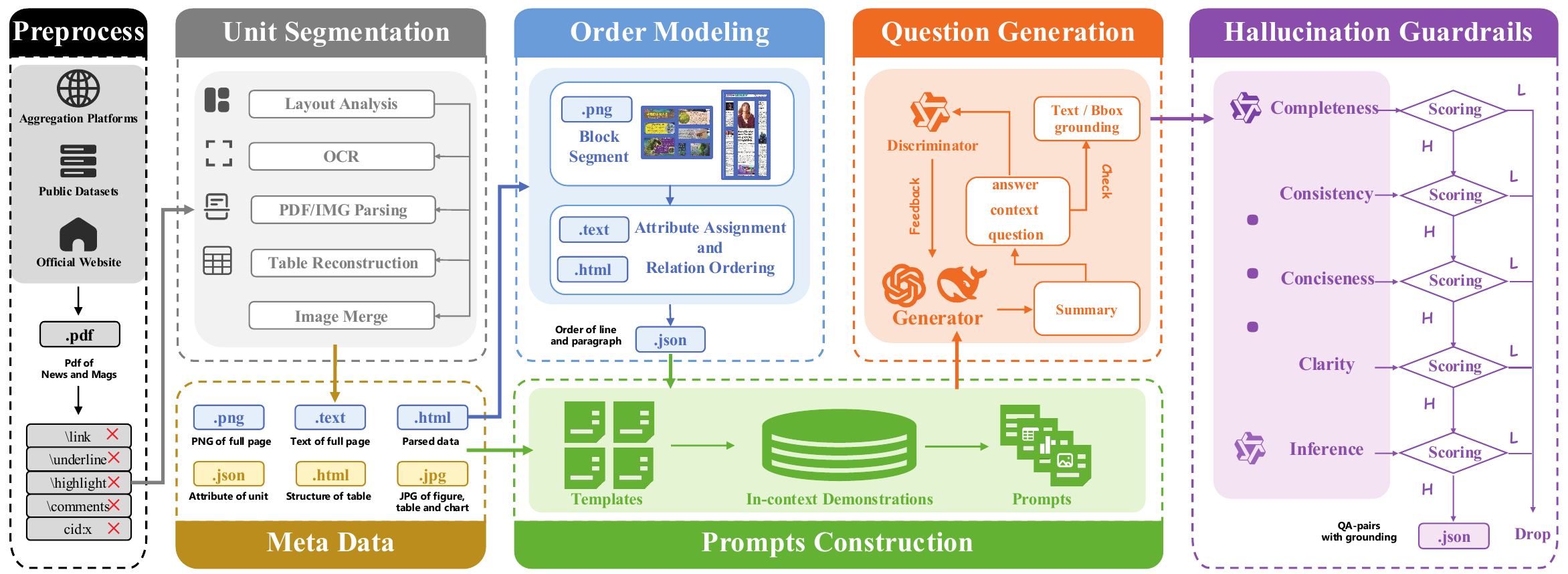}
\caption{Overview of DocWeaver, a Multi-Agent Pipeline for Generating the MosaicDoc Benchmark.}
\label{fig:docweaver}
\end{figure*}

\section{Related Work}
We review key datasets in Document Visual Understanding (DVU), focusing on two core tasks where current benchmarks show significant limitations: Reading Order Prediction (ROP) and Document Visual Question Answering (DocVQA).

\subsection{Datasets for Reading Order Prediction}
Reading Order Prediction, which aims to infer coherent reading sequences from 2D layouts, is a cornerstone of document comprehension. While methods like LayoutReader~\cite{wang2021layoutreader} and Dolphin~\cite{feng2025dolphin} have drawn attention, progress is constrained by the lack of robust and diverse datasets. Early examples like ReadingBank~\cite{wang2021layoutreader} provides word-level text sequences for 500K documents but lacks block-level annotations for structural evaluation, which is critical for complex layouts like multi-column Manhattan~\cite{liu2025xy-cut}. Other efforts, such as ROOR~\cite{zhang2024roor}, augments existing datasets like FUNSD~\cite{jaume2019funsd} with reading order but remains limited in scale and layout diversity.

More recently, large-scale ROP datasets have emerged, including DocGenome~\cite{xia2024docgenome} on scientific papers and olmOCR-mix-0225~\cite{poznanski2025olmocr} on web-scraped PDFs. However, a critical limitation persists across all these benchmarks: they are dominated by documents with simple, homogeneous layouts(e.g., single- or two-column academic papers). This fails to capture the complexity of real-world documents with non-Manhattan or multi-column layouts, thereby providing an inadequate challenge for evaluating robust reading order prediction. Furthermore, they are often single-task and mono-lingual.

\subsection{Datasets for Document Visual Question Answering}
DocVQA benchmark similar lack real-world complexity. Many datasets target individual pages or components, with multi-page understanding being a nascent and flawed area. For instance, MP-DocVQA~\cite{tito2023mpdocvqa} extends SP-DocVQA~\cite{docvqa} but lacks questions that require reasoning across pages, while DUDE~\cite{van2023dude} suffers from short contextual spans and inconsistent quality due to crowdsourcing. Other datasets are domain-specific, such as TAT-DQA~\cite{zhu2024tatdqa} for financial reports emphasizing structured tables and numerical reasoning within a narrow domain, SlideVQA (Tanaka et al. 2023) for low-density presentation slides, and InfographicsVQA~\cite{mathew2022infographicvqa}, which prioritizes visual elements like charts and icons but underrepresents dense text and complex layouts. A crucial deficiency across nearly all these datasets is the scarcity of questions requiring multi-span reasoning.

Recent LLM-based generation methods, such as RefChartQA~\cite{vogel2025refchartqa} for charts and TVG~\cite{liu2024tvg} for tables, primarily automate annotation for well-structured elements rather than introducing new, complex visual domains. Even TRIG~\cite{li2025TRIG} provides visual grounding, ultimately falls back on human validation, failing to establish a fully automated, scalable pipeline.

Our work directly addresses these identified gaps. The DocWeaver pipeline introduces a fully automated generation and validation process, moving beyond the manual verification required by prior work. Most critically, the resulting MosaicDoc benchmark introduces visually complex and stylistically diverse newspaper and magazine documents in both English and Chinese. By providing rich, multi-task annotations for this challenging new domain, MosaicDoc fills a crucial void in the DVU landscape, offering a more realistic and rigorous testbed for future research.

\section{The DocWeaver Pipeline}
To automate the creation of a large-scale, multi-task benchmark from visually rich document, we introduce DocWeaver, a multi-agent collaborative pipeline. As illustrated in Figure~\ref{fig:docweaver}, DocWeaver is designed to systematically decompose complex documents, generate high-quality questions, and rigorously validate the outputs without manual intervention. The pipeline operates in three phases: (1) Document Decomposition and Structuring, (2) High-Fidelity QA Generation, and (3) Automated Quality Assurance.

\subsection{Document Decomposition and Structuring}
This initial phase transforms raw PDF documents into a structured, machine-readable format.

\noindent\textbf{Data Preprocessing}\hspace{1.3em}The pipeline parses digital PDFs using PyMuPDF\textsuperscript{1} and pdfminer.six\textsuperscript{2} to extract raw text and bounding boxes. Semantically irrelevant elements such as annotations, hyperlinks, underlines, and highlights are discarded. To handle encoding errors in PDFs(e.g., malformed symbols like \textless?\textgreater{} or cid:x), we employ chardet\textsuperscript{3} to detect encoding, ensuring text recovery. Each page is rendered into two raster images: one at high resolution for precise layout analysis, and another at randomized DPI to simulate real-world rendering variance and increase dataset diversity.

\footnotetext[1]{https://pymupdf.readthedocs.io}
\footnotetext[2]{https://pdfminersix.readthedocs.io/}
\footnotetext[3]{https://pypi.org/project/chardet/}

\noindent\textbf{Unit Segmentation}\hspace{1.3em}We employ a suite of specialized agents to decompose each document page into its constituent semantic units. The process begins with a Layout Detection Agent, PP-DocLayout~\cite{sun2025pplayout} fine-tuned on the $\text{M}^6\text{Doc}$ magazine dataset~\cite{cheng2023m6doc}, which identifies 13 distinct layout element types (e.g., titles, paragraphs, and subhead). To recover text missed by PDF parsers, a Multi-Engine OCR Agent then processes each detected layout element with three different OCR engines\textsuperscript{4}. The final text is determined via a weighted-voting scheme that considers both engine reliability and confidence scores.

\footnotetext[4]{\url{https://github.com/PaddlePaddle/PaddleOCR}\newline
\url{https://pypi.org/project/pytesseract/}\newline
\url{https://github.com/JaidedAI/EasyOCR}}

\begin{equation}
\label{eq:example}
\hat{s} = \arg\max_{t \in \mathcal{T}} \sum_{i = 1}^{N} w_i p_i \mathbf{1}\!\left[s_i = t\right]
\end{equation}
where $\mathcal{T}=\{s_1,s_2,...s_N\}$ denotes all candidate strings from each OCR engine. $\mathbf 1[\cdot]$ is the indicator function, $w_i$ is the weight reflecting the expected reliability of the $i$-th engine, and $p_i$ is the confidence score. Following recognition, global char-level positions are recovered from PaddleOCR's CTC decoder.

 Subsequently, specialized agents handle structural and relational information. A Table Structure Agent, using the TableStructureRec toolbox\textsuperscript{5}, converts the detected tables to a structured HTML format. Concurrently, a Reading Order Agent powered by MinerU~\cite{wang2024mineru} predicts an initial reading order for documents with standard layouts. To handle content spanning multiple pages, a Cross-Page Linking Agent leverages DeepSeek\textsuperscript{6} to compute semantic similarity between adjacent pages. Pages with a similarity score exceeding 0.8 are then automatically merged into a single logical document, up to a four pages limit. This structured decomposition provides a high-fidenlity foundation for all subsequent annotation tasks.

\footnotetext[5]{https://github.com/RapidAI/TableStructureRec}
\footnotetext[6]{https://www.deepseek.com/}

\begin{table*}[t]
\centering
\small
\setlength{\tabcolsep}{10pt} 
\renewcommand{\arraystretch}{1.2} 
\begin{tabular}{c|ccccccc}
\toprule
\rule{0pt}{12pt}\textbf{Dataset}                      & \textbf{Sources}       & \textbf{\#Images} & \textbf{Tokens} \textit{(avg$\pm$std )}         & \textbf{\#Tasks} & \textbf{Lang} & \textbf{\#Ques} & \textbf{Unique} \textit{(\%)}  \\[3pt]
\toprule

\multicolumn{8}{l}{\hspace{0.1em}\scriptsize\textit{Document Visual Question Answering (DocVQA)}} \\

\midrule
MP-DocVQA         & Industry docs & 12.7K   & $489.6_{\pm 411.2}$       & 2     & en     & 41.4K  & 84.2      \\
TAT-DQA                & Finance reports & 2.8K    & $850.9_{\pm 252.0}$       & 2     & en     & 16.6K  & 88.0     \\
InfographicsVQA       & Infographics & 5.5K    & $382.7_{\pm 231.7}$       & 3     & en     & 27.1K  & 98.5       \\
DuReader$_{vis}$        & Web           & 12.6K   & $3,186.8_{\pm 1,902.3}$   & 2     & zh     & 14.1K  & 99.7      \\
DUDE                  & Multi         & 27.9K   & $2,944.0_{\pm 3,937.1}$   & 4     & en     & 30.1K  & 86.9      \\
\midrule
\multicolumn{8}{l}{\hspace{0.1em}\scriptsize\textit{Reading Order Prediction (ROP)}} \\
\midrule
ROOR                  & Scan forms    & 0.2K    & $323.6_{\pm 149.3}$       & 3     & en     &  -     & -      \\
ReadingBank           & Ebooks        & 500K    & $314.6_{\pm 144.6}$       & 2     & en     &  -     & -      \\
olmOCR-mix-0225       & Web, Ebooks   & 266K    & $661.0_{\pm 598.7}$       & 1     & en     &  -     & -      \\
\midrule

\midrule
\multirow{2}{*}{\textbf{MosaicDoc} (Ours)}       & Magzines      & 42.7K   & $1,075.1_{\pm 860.6}$     & 6     & en, zh & 304.6K  & 99.7      \\
       & Newspapers    & 29.6K   & $3,557.9_{\pm2051.2}$     & 6     & en, zh & 318.5K  & 99.8      \\ 
\bottomrule
\end{tabular}
\caption{Comparison of MosaicDoc with existing benchmarks for Document Visual Understanding. MosaicDoc is the first large-scale benchmark to combine visually rich sources (magazines, newspapers), bilingual support (English and Chinese), and comprehensive multi-task annotations. This directly addresses the key limitations of prior work, which are often restricted to simpler layouts, a single language, or fewer tasks.}
\label{tab: benchmarks}
\end{table*}

\begin{figure*}[t]
\centering
\includegraphics[width=0.9\textwidth]{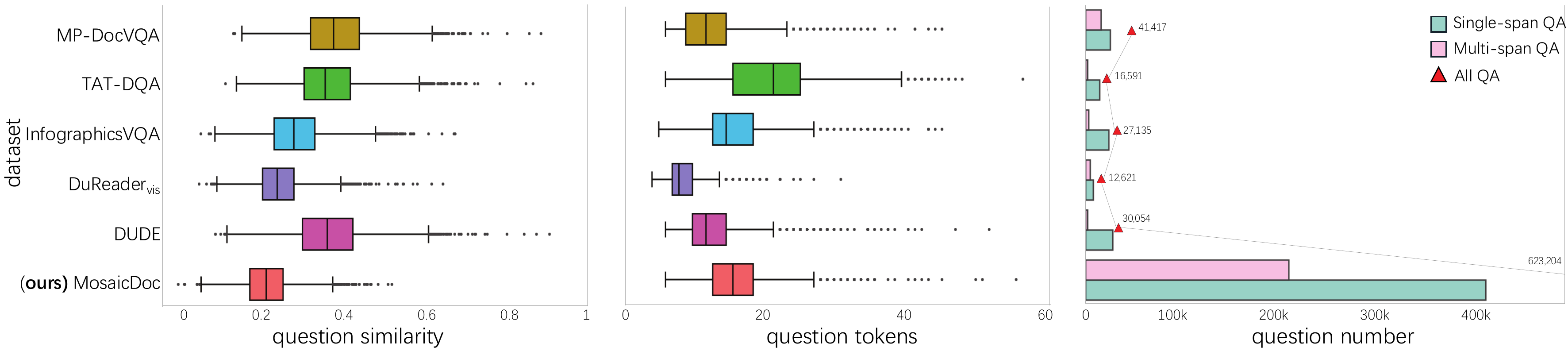}
\caption{Left panels show the distributions of question similarity and token length, while right panel compares number of multi-span and single-span question-answer pairs instances}
\label{fig:question-aspect}
\end{figure*}

\noindent\textbf{Complex Reading Order Modeling}\hspace{1.3em}A key challenge in VRDU is modeling the non-linear and diverse reading orders often found in newspapers and magazines. Our approach addresses this through two tailored strategies (Please refer to Appendix A for more details).

For Documents with Structured Data (e.g., magazines and Chinese newspapers) where official HTML versions are available, we use the HTML structure as a reference template. Text spans extracted from the PDF are mapped to the template using fuzzy matching with edit distance, constrained by bounding box and contextual text proximity. The final correct reading order is then derived from the validated sequence within each layout block.

For Documents without Structured Data (e.g., English newspapers), we employ a hybrid approach combining structural analysis and semantic clustering. Page images are first binarized to detect layout lines, which are used to partition the page into rectangular content blocks, each representing a distinct article. Then, within each block, line-level reading order is inferred using PDF metadata (e.g., font features, positional cues) and semantic coherence.

\subsection{High-Fidelity QA Generation}

This phase leverages the structured document representation to generate diverse and challenging question-answer pairs.

\noindent\textbf{Prompts Construction}\hspace{1.3em}To elicit a wide range of question types, we developed specialized prompt templates for text, tables, and charts (see Appendix B for details). These templates guide the model to generate a diverse question set. Additionally, we build an in-context learning repository containing examples for single-span, multi-span, and table- and chart-based QA. We source examples from datasets such as MultiSpanQA\cite{li2022multispanqa}, InfographicsVQA\cite{mathew2022infographicvqa} and TableBench\cite{wu2025tablebench}, supplemented with 400 manually crafted examples for domain-specific coverage. During generation, we randomly sample four well-designed and four dataset-sourced examples to serve as few-shot demonstrations within the prompt.

\noindent\textbf{Question Generation}\hspace{1.3em}We employ powerful LLMs, GPT-4o\textsuperscript{7}\footnotetext[7]{\url{https://openai.com/es-ES/api/}} and DeepSeek-R1, as QA generators. To ensure high quality and factual grounding, the generation process follows a structured, multi-step workflow. First, the model generates a summary of the input content to identify key information. Based on the summary, a question is formulated. Then, the model is explicitly instructed to locate the corresponding answer-bearing context within the original content and to extract the answer directly, avoiding reliance on prior knowledge. Furthermore, we introduce QWQ-32B \cite{team2025qwq} as an auxiliary discriminator that assesses whether the retrieved context supports cross-sentence reasoning. If so, the context along with a prompt is returned to the generator with explicit instructions to formulate multi-span questions with detailed reasoning chains and supporting evidence. The final answer and its evidence are then matched back to precisely locate their positions in the document image.

\subsection{Automated Quality Assurance}

\noindent\textbf{Hallucination Guardrails}\hspace{1.3em}Despite rigorous prompting, minor hallucinations may still occur. Inspired by the G-EVAL\cite{liu2023g-eval} framework, we implement a final filtering stage using QWQ as an LLM evaluator serving as an automated "hallucination guardrail". Each generated QA pair is scored from 1 to 5 across five distinct criteria, and only those passing a high threshold on all criteria are retained (see Appendix B for details). The five evaluation criteria are:
\begin{itemize}
\item \textbf{Completeness:} Assess if the answer contains all essential elements without omitting any critical information.
\item \textbf{Consistency:} Verify the answer factually aligns with source document.
\item \textbf{Conciseness:} Evaluate if the answer avoids redundancy information.
\item \textbf{Clarity:} Determine if the answer is free from ambiguous references (e.g., "this year", "here").
\item \textbf{inference:} Validates the precision of any calculations or logical reasoning.
\end{itemize}

\section{The MosaicDoc Benchmark}

This section details the construction, quality assurance process, and key statistical properties of the MosaicDoc benchmark.

\subsection{Data Sources}
\begin{figure}[t]
\centering
\includegraphics[width=0.3\textwidth]{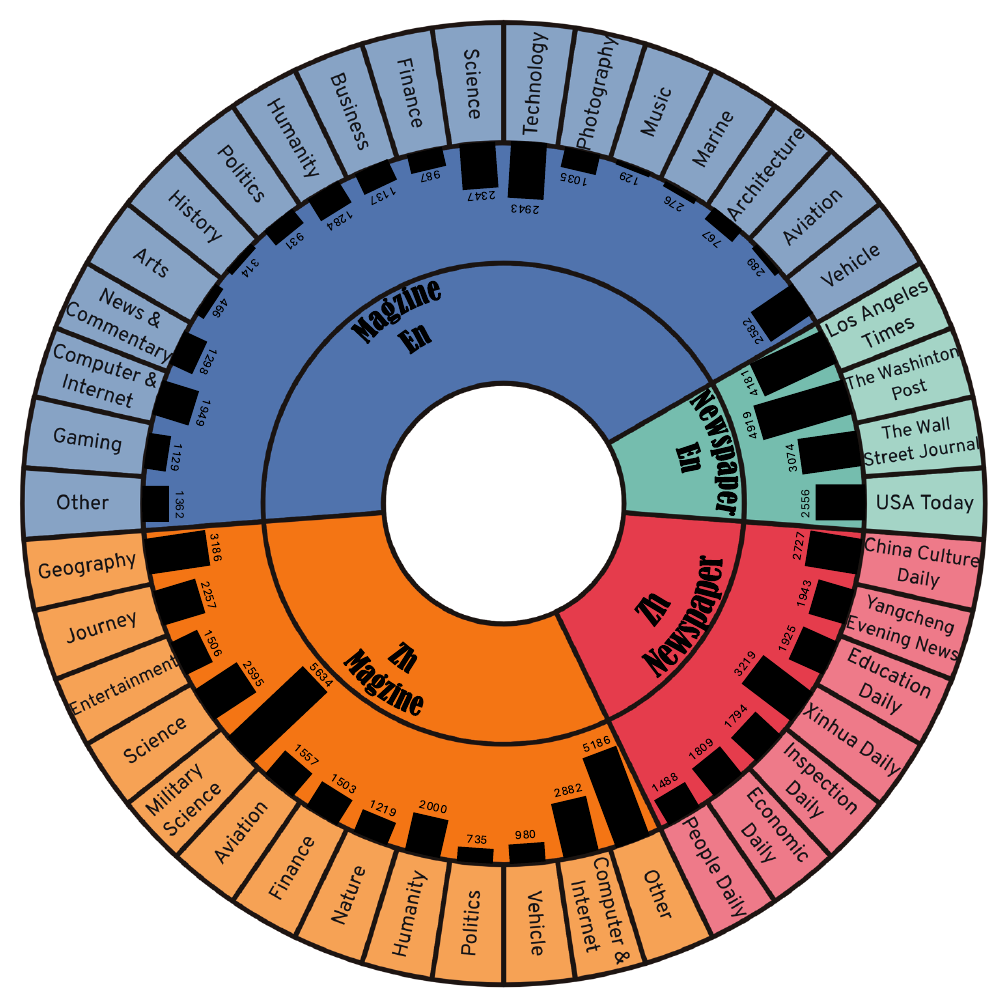}
\caption{Source category distribution of MosaicDoc dataset.}
\label{fig:categories}
\end{figure}

To construct a benchmark that reflects real-world visual diversity, we carefully curated a collection of documents from official public websites, content-rich aggregation platforms, and open-source PDF repositories, ensuring compliance with all data usage and licensing requirements (details in Appendix D). The resulting MosaicDoc benchmark is composed of four subsets: Chinese magazines, Chinese newspapers, English magazines, and English newspapers. The data is sourced from 196 distinct publishing institutions across 24 diverse domains, including science, finance, culture, and the arts. The full distribution of sources is visualized in Figure~\ref{fig:categories}.

\subsection{Quality Assurance}
\textbf{Automated Filtering}\hspace{1.3em}As described in the DocWeaver section, every generated question-answer pair is passed through our five-criteria "Hallucination Guardrail" system, with only high-scoring pairs being retained (results in Table~\ref{fig:filter}). Additionally, we apply a similar automated check for reading order. If a document page contains more than five detected reading order errors judged by distance-based heuristics (see Appendix C), the entire page is discarded.

\noindent\textbf{Human Validation}\hspace{1.3em}To establish a gold-standard test set for evaluation, we conducted a meticulous manual validation. We randomly sampled 200 document images and their corresponding annotations from across all four subsets. For this sample, human annotators manually corrected any remaining errors in the QA pairs and reading order sequences, following the same criteria as our automated guardrails. This high-fidelity, human-validated subset is used for all experimental evaluations reported in this paper, ensuring our results are grounded in the most accurate data possible.

\begin{table}[t]
\centering
\small
\setlength{\tabcolsep}{4pt} 
\renewcommand{\arraystretch}{1.2} 
\tabcolsep=7pt
\begin{tabular}{cccc}
\toprule
& \textbf{Previous} & \textbf{After} & \textbf{Ratio} \textit{(\%)} \\ 
\hline
en-Magzine     & 178.2k & 160.1k  & 89.8 \\
zh-Magzine     & 164.7k & 144.5k  & 87.7  \\
en-Newspaper   & 184.7k & 163.6k  & 88.6  \\
zh-Newspaper   & 176.1k & 154.9k  & 87.9 \\
\bottomrule
\end{tabular}
\caption{The counts of QA pairs after guardrails filtering}
\label{fig:filter}
\end{table}

\begin{table*}[ht]
\centering
\scriptsize
\setlength{\tabcolsep}{5pt} 
\begin{tabular}{cccccccccccc}
\toprule
\multirow{2}{*}{\textbf{Sou.}}  & \multirow{2}{*}{\textbf{Type}} & \multicolumn{2}{c}{\textbf{Expert Models}} & \multicolumn{3}{c}{\textbf{Expert VLMs}}  & \multicolumn{5}{c}{\textbf{General VLMs}} \\
\cmidrule(l{3pt}r{3pt}){3-4} \cmidrule(l{10pt}r{20pt}){5-7}  \cmidrule(l{90pt}r{5pt}){7-12}
                      &  & \textbf{Donut}    & \textbf{ViTLP}    & \textbf{Vary}    & \textbf{TextMonkey} & \textbf{mPlug-DocOWL2}  & \textbf{CogVLM2}     & \textbf{InternVL3}    & \textbf{Qwen2.5-VL} & \textbf{GPT-4o} & \textbf{Gemini-2.5 } \\ 
\hline
\hline
\multicolumn{2}{c}{Parameters} & 253M & 259M & 7B & 7B & 7B & 19B & 9B & 7B  & API  & API \\
\hline

\multirow{5}{*}{Mag.} & S     & 8.13 / --    & 6.91 / --    & 1.78 / --  & 5.43 / --    & 13.75 / 3.11  & 31.86 / 18.79 & 54.11 / 48.68  & 57.86 / 58.59 & 47.64 / 18.76 & 65.78 / 59.27 \\ 
                      & M     & 1.26 / --    & 1.28 / --    & 0.49 / --  & 4.65 / --    & 5.38 / 5.73   & 23.77 / 18.38 & 40.78 / 38.68  & 43.81 / 40.39 & 42.08 / 18.03 & 55.63 / 63.67 \\ 
                      & T     & 13.11 / --   & 10.04 / --   & 8.4 / --   & 3.02 / --    & 18.54 / 8.27  & 30.42 / 19.04 & 53.53 / 45.27  & 54.62 / 54.84 & 46.19 / 28.06 & 68.37 / 70.64 \\ 
                      & C     & 3.38 / --    & 5.78 / --    & 2.18 / --  & 1.38 / --    & 22.72 / 14.95 & 30.46 / 15.71 & 33.42 / 33.32  & 41.42 / 41.88 & 32.80 / 32.69 & 42.65 / 50.39 \\ 
                      & All   & 5.87  / --   & 5.14 / --    & 1.81 / --  & 4.78 / --    & 11.70 / 3.97  & 28.93 / 18.41 & 48.43 / 43.98  & 51.98 / 51.33 & 48.86 / 20.75 & 61.27 / 68.82 \\ 
\hline
\multirow{5}{*}{News.}  & S     & 4.54  / --    & 3.80 / --    & 3.87 / --  & 2.16 / --    & 9.80 / 1.62    & 18.43 / 7.98  & 51.81 / 38.44  & 61.19 / 51.63 & 32.34 / 10.70 & 68.47 /                           59.05 \\ 
                        & M     & 0.50  / --    & 0.68 / --    & 0.62 / --  & 2.86 / --    & 1.45 / 0.04    & 11.50  / 2.10 & 29.26 / 26.54  & 36.42 / 35.71 & 25.14 / 6.55 & 55.14 / 53.97 \\ 
                        & T     & 2.62  / --    & 0.79 / --    & 6.15 / --  & 0.00 / --    & 6.12 / 2.73    & 8.82 / 10.90  & 39.05 / 33.59  & 41.37 / 50.02 & 21.42 / 13.74 & 48.55 / 75.16 \\ 
                        & C     & 5.72  / --    & 3.52 / --    & 7.69 / --  & 5.43 / --    & 27.10 / 16.53  & 22.09 / 28.82 & 39.09 / 34.74  & 42.35 / 43.67 & 29.57 / 29.80 & 48.22 / 62.91 \\ 
                        & All   & 3.63  / --    & 7.69 / --    & 3.70 / --  & 2.32 / --    & 8.97 / 1.98    & 16.37 / 6.65  & 45.80 / 31.77  & 52.99 / 44.79 & 29.65 / 10.17 & 62.47 / 59.76 \\ 
\bottomrule
\end{tabular}
\caption{The ANLSL score of Baseline performance on the MosaicDoc dataset. The figures before the "/" denote English subset, while those after it denote Chinese subset. The type of questions are abbreviated as (S)ingle-span, (M)ulti-span, (T)able and (C)hart. Sou. means source, Mag. means magzine and News. means newspaper. }
\label{tab:docvqa}
\end{table*}

\subsection{Dataset Properties and Statistics}

MosaicDoc was created to fill a critical resource gap in Document AI research. Its properties demonstrate its significant advantages over existing datasets in scale, complexity, and task diversity.

\noindent\textbf{Scale and Composition}\hspace{1.3em}The full benchmark comprises over 72.3K document images from newspapers and magazines, annotated with fine-grained OCR and reading order labels. Layered on top are over 620K question-answer pairs that query content from text, tables and charts.

\begin{figure}[t]
\centering
\includegraphics[width=0.5\textwidth]{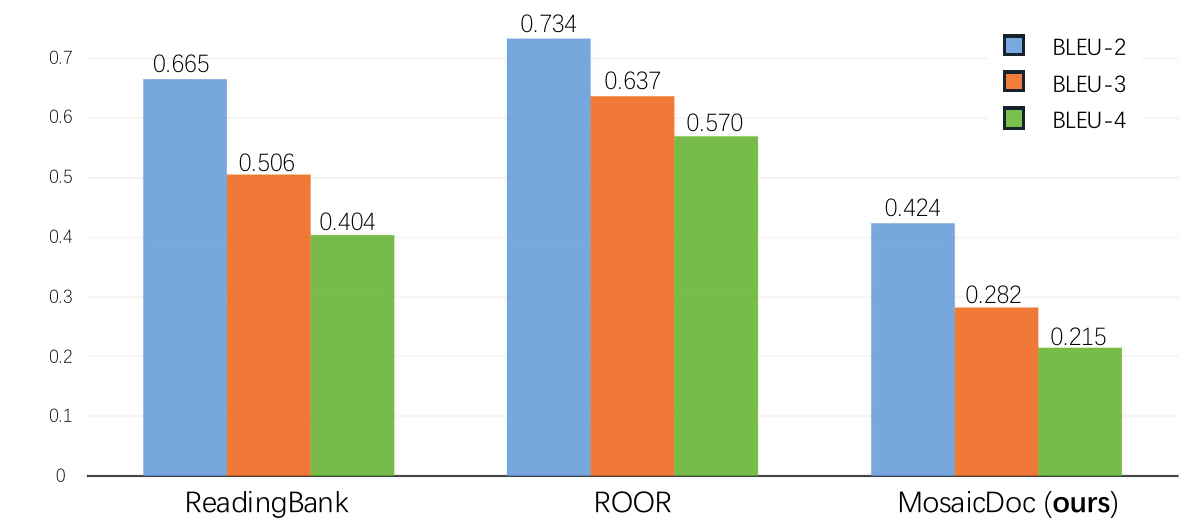}
\caption{The BLEU scores are calculated for the left-to-right and top-to-bottom order to measure the layout complexity of MosaicDoc.}
\label{fig:bleu-bechmark}
\end{figure}

\noindent\textbf{Textual and Layout Complexity}\hspace{1.3em}MosaicDoc is defined by its information density. As shown in Table~\ref{tab: benchmarks}, the average number of tokens per document is significantly higher than in most other VQA datasets. While DuReader-vis also features high token counts, its web-page screenshots often contain sparse layouts where answers are confined to small regions. In contrast, MosaicDoc's sources feature dense, structured text across the entire page. The complexity of these layouts is further quantified by average sentence-level BLEU metrics (see Figure~\ref{fig:bleu-bechmark}), confirming that MosaicDoc presents a more challenging structural reasoning task.

\noindent\textbf{Question Quality and Diversity}\hspace{1.3em}The questions generated by DocWeaver are not only numerous but also sophisticated. Using a suite of Qwen models\textsuperscript{8},\footnotetext[8]{\url{https://huggingface.co/Qwen/Qwen3-Embedding-0.6B}} we sample and analyze 5,000 questions per dataset, computing semantic similarity scores and token lengths. We found that questions in MosaicDoc have a higher average token length and significantly lower semantic similarity to each other compared to other datasets, indicating greater diversity and less repetition. With a uniqueness score of over 99.7\%, nearly every question is distinct. Crucially, MosaicDoc contains a substantial volume of multi-span QA pairs, addressing a well-known gap in existing DocVQA datasets and enabling the evaluation of complex answer extraction scenarios that require aggregating information from multiple spans within a document.

\noindent\textbf{Comprehensive Multi-Task Support }\hspace{1.3em}Unlike single-task benchmarks like ReadingBank or domain-specific ones like DuReader-vis, MosaicDoc provides a unified platform for a broad range of VRDU tasks. Its rich annotations support: (1) Document VQA, (2) Word- and line-level OCR, (3) Block- and line-level Reading Order Prediction, and (4) Content-aware Localization. By providing a single, comprehensive resource that bridges these tasks, MosaicDoc serves as a more holistic and challenging benchmark for advancing Document AI.

\section{Evaluation}

To demonstrate the challenges posed by MosaicDoc and to establish a new performance baseline for Visually Rich Document Understanding (VRDU), we conduct a comprehensive evaluation of 13 SOTA models.

\subsection{Experimental Setting}
\noindent\textbf{Baselines}\hspace{1.3em}
We categorize the 13 recent SOTA models into three groups:
\begin{itemize}
\item \textbf{Expert Models} include pre-LLM architectures like LayoutReader\cite{wang2021layoutreader}, Donut\cite{kim2022donut}, and ViTLP\cite{mao2024vitlp}. LayoutReader uses line-level text bounding boxes as input, whereas Donut and ViTLP consume only the image and question. None of these models incorporate large language models (LLMs), and all have smaller parameter counts ($<$0.3B).

\item \textbf{Expert VLMs} comprise models like Vary-7B\cite{wei2024vary}, mPLUG-DocOwl2-7B\cite{hu2024mplug-docowl2}, TextMonkey-7B\cite{liu2024textmonkey}, GOT-OCR-0.5B\cite{wei2024got-ocr}, and olmOCR-8B\cite{poznanski2025olmocr}. These  are built on LLM backbones and are specifically pretrained on large document datasets for tasks like DocVQA and ROP.

\item \textbf{General VLMs} include powerful, general-purpose vision-language models like CogVLM2-19B\cite{hong2024cogvlm2}, InternVL3-9B\cite{zhu2025internvl3}, and Qwen2.5-VL-7B\cite{bai2025qwen2.5}. In addition, we accessed GPT-4o and Gemini-2.5\textsuperscript{9}\cite{comanici2025gemini} through their APIs. While not specialized for documents, they possess state-of-the-art capabilities across a wide range of vision-language tasks.
\end{itemize}

\footnotetext[9]{\url{https://gemini.google.com/}}

\noindent\textbf{Implementation Details}\hspace{1.3em}
All models are evaluated in a zero-shot setting using their official default configurations. Images are provided at the maximum resolution supported by each model. All VLM evaluations were conducted within the VLLM framework on NVIDIA A100 GPUs with 80GB memory. Further implementation details and qualitative results are provided in Appendix E.

\subsection{Results and Analysis}
We evaluate model performance on three core VRDU tasks: Document VQA, Page-Level OCR, and Reading Order Prediction.

\noindent\textbf{Document VQA}\hspace{1.3em}We evaluate DocVQA performance using the Average Normalized Levenshtein Similarity for List (ANLSL) metric (~\cite{tito2021doccvqa}). The comprehensive results, broken down by question type, are presented in Table 3.

Our analysis reveals several key findings. First, General VLMs significantly outperform both Expert Models and a majority of the specialized Expert VLMs. This suggests that the massive, diverse pre-training data and superior architectural scaling of general models provide a more robust foundation for VRDU than domain-specific pre-training on simpler documents.

Second, Expert VLMs struggle unexpectedly on MosaicDoc's sources. We primarily attribute this to token reduction strategies employed by models like TextMonkey and mPLUG-DocOwl2. While effective for simple documents, these methods, which merge or discard visual tokens, likely cause critical semantic loss when applied to the information-dense layouts of magazines and newspapers.

Third, all models exhibit a dramatic and consistent performance drop on multi-span questions. Even the top-performing local model Qwen2.5-VL and the remote model Gemini-2.5 see a sharp decline in ANLS score compared to their single-span performance. This highlights a critical and universal weakness in current models' ability to perform multi-span extraction that requires synthesizing information across complex layouts. Interestingly, performance on table and chart questions is comparatively higher, likely because these tasks rely more on locating structured objects than on fine-grained reading of dense paragraphs.

\noindent\textbf{Page-level OCR}\hspace{1.3em}To assess raw text recognition capability, we evaluate page-level Character Recognition Rate (OCR) using CRR and Output-based Character Recognition Rate (OCRR). CRR measures accuracy against the ground truth, while OCRR normalizes by the length of the model's output, measuring precision within the generated text.

As shown in Table~\ref{tab:ocr}, all models perform significantly worse on the newspaper subset, particularly for Chinese newspapers, underscoring the extreme difficulty of their dense and complex layouts. We also observe a widespread and common failure mode where VLMs, after generating a certain amount of text, begin producing repetitive sequences until reaching their token limit, resulting in lower OCRR. This severely penalizes the CRR score and indicates a breakdown in contextual understanding when processing long, dense visual inputs, even when the initial recognized text is accurate. This failure to read the full page effectively limits a model's ability to answer content-related questions.

\noindent\textbf{Reading Order Prediction}\hspace{1.3em}We evaluate ROP using the Micro-F1 score on text line sequences. Since VLMs produce plain text, we determine the correctness of the predicted order by matching text blocks to the ground truth sequence.

The results in Table~\ref{tab:line_rop} reveal a highly consistent trend across most models: high precision paired with low recall. This indicates that while the text fragments models do recognize are often in the correct local sequence (e.g., within a single column), they fail to capture the entire content of the page. This problem is exacerbated in newspapers, where horizontally adjacent blocks are often spatially closer than vertically sequential lines within an article, challenging simple top-down heuristics. Despite strong performance across the three subsets, Gemini can only effectively order text within the same column and fails to establish the correct sequential relationship between paragraph-level texts in multi-column or non-Manhattan layouts—a challenge that lies at the core of reading order recovery in such complex document layouts (see results and visualizations in Appendix F.4). This failure to comprehend the global layout contributes to the repetitive outputs seen in the OCR task and fundamentally undermines the model's ability to understand the document as a whole.

\begin{table}[t]
\centering
\scriptsize
\setlength{\tabcolsep}{4pt} 
\renewcommand{\arraystretch}{1.2} 
\begin{tabular}{ccccccc}
\toprule
\multirow{2}{*}{\textbf{Model}}  & \multicolumn{2}{c}{\textbf{Magzine} (en / zh)} & \multicolumn{2}{c}{\textbf{Newspaper} (en / zh)}  \\
\cmidrule(l{5pt}r{6pt}){2-3} \cmidrule(l{5pt}r{6pt}){4-5}
                      & $\mathbf{CRR}\!\uparrow$ & $\mathbf{OCRR}\!\uparrow$
                      & $\mathbf{CRR}\!\uparrow$     & $\mathbf{OCRR}\!\uparrow$ \\
\hline  
GOT-OCR          & 22.27 / 31.46     & 45.62 / 41.15     & 3.09 / 7.15     & 1.13 / 5.17     \\
olmOCR           & 73.76 / 53.85    & 86.33 / 74.20    & 32.58 / 1.19  & 52.00 / 12.45    \\
InternVL3         & 66.06 / 59.29    & 74.65 / 61.38    & 56.03 / 31.77  & 58.59 / 44.89    \\
Qwen2.5-VL       & 55.24 / 57.29   & 23.60 / 28.40    & 34.46 / 39.62 & 11.17 / 31.40    \\
GPT-4o       & 69.81 / 30.33   & 75.88 / 39.15    & 37.55 / 2.69  & 53.17 / 15.34    \\
Gemini-2.5       & 89.42 / 87.34   & 90.32 / 80.31    & 87.90 / 66.64  & 91.04 / 78.18    \\
\bottomrule
\end{tabular}
\caption{Page-level OCR recognition results on MosaicDoc}
\label{tab:ocr}
\end{table}

\begin{table}[t]
\centering
\scriptsize
\setlength{\tabcolsep}{2.5pt} 
\renewcommand{\arraystretch}{1.2} 
\begin{tabular}{ccccccc}
\toprule
\multirow{2}{*}{\textbf{Model}}  & \multicolumn{3}{c}{\textbf{Magzine} (en / zh)} & \multicolumn{3}{c}{\textbf{Newspaper} (en / zh)}  \\
\cmidrule(l{5pt}r{6pt}){2-4} \cmidrule(l{5pt}r{6pt}){5-7}
                      & $\mathbf{P}$    & $\mathbf{R}$    & $\mathbf{F1}$    & $\mathbf{P}$ & $\mathbf{R}$  & $\mathbf{F1}$ \\ 
\hline
LayoutReader     & 5.93/10.6    & -- / --      & -- / --    & 5.40/3.19    & -- / --    & -- / --   \\
GOT-OCR          & 2.55/7.05     & 0.61/2.72      & 0.99/3.93    & 14.2/13.5    & 0.55/0.17  & 1.07/0.34 \\
olmOCR           & 92.7/91.1    & 73.3/58.7   & 81.9/71.4  & 60.1/22.6   & 23.5/0.36  & 33.8/0.71 \\
InternVL3         & 86.1/87.6    & 62.1/70.7    & 72.2/76.6  & 82.2/74.8    & 52.2/28.4  & 63.9/41.2 \\
Qwen2.5-VL       & 92.6/93.5    & 50.6/63.6    & 65.4/75.7  & 90.0/92.0    & 35.5/41.5  & 50.9/57.2 \\
GPT-4o       & 85.9/74.6    & 60.1/33.2    & 70.7/45.9  & 74.6/60.0   & 33.2/17.3  & 45.1/26.9 \\
Gemini-2.5       & 91.9/93.8    & 84.2/85.4    & 87.9/89.4  & 95.7/94.2   & 81.3/40.2  & 87.9/56.3 \\
\bottomrule
\end{tabular}
\caption{Text Line reading order prediction results}
\label{tab:line_rop}
\end{table}

\section{Conclusion}
In this work, we introduced DocWeaver, a novel multi-agent pipeline for automated data generation, and used it to construct MosaicDoc, a large-scale bilingual benchmark for VRDU. By focusing on complex newspaper and magazine documents (a domain largely neglected by previous research), MosaicDoc provides the community with a more realistic and challenging testbed. Its diversity in language, layout, and task support addresses critical limitations of existing datasets. Our extensive evaluation of state-of-the-art models on MosaicDoc reveals significant and previously unmeasured systemic weaknesses, particularly in handling dense layouts and performing multi-span reasoning, thereby charting a clear path for future research. While DocWeaver has proven effective, we view this as a foundational step. Future work will focus on extending the pipeline to diverse new domains, such as historical and handwritten documents, to progressively further push the boundaries of robust document intelligence.

\bibliography{aaai2026}

\appendix                   
\clearpage 
\section*{Appendix}
This appendix provides supplementary information omitted from the main paper due to space limitations. Specifically, Section A elaborates on the detailed processes for document information extraction and reading order construction. Section B discusses the implementation details of DocWeaver. Section C covers the data curation operations for parts of the MosaicDoc dataset, while Section D presents additional properties of the dataset. Section E provides the mathematical formulas for the metrics used in our experiments. Finally, Section F showcases qualitative results on the MosaicDoc dataset.


\section{A. Details for Document Structuring}
This section elaborates on the procedures for document metadata extraction and structural layout construction, covering layout analysis, text block segmentation, and reading order determination.

\subsection{A.1 Layout Analysis for Magzine}
We applied for the entire $M^6$ dataset~\cite{cheng2023m6doc} and re-divided the magazine subset into 13 layout elements, as shown in Figure 6. Subsequently, PP-DocLayout-L was fine-tuned on this subset for 100 epochs, improving the precision from an initial 0.583 to a final 0.871.

\begin{figure}[ht]
\centering
\includegraphics[width=0.45\textwidth]{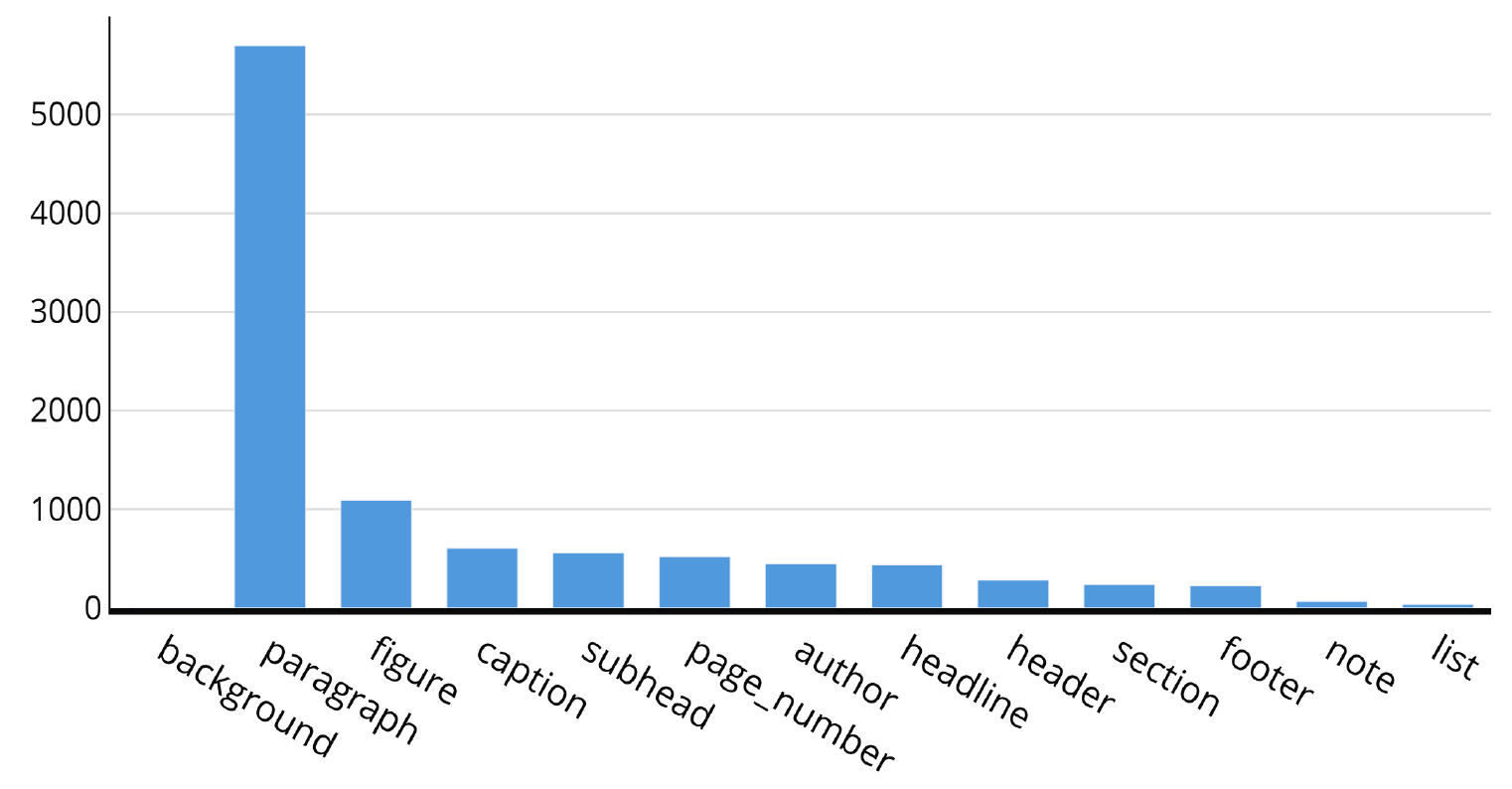}
\caption{The distribution of 13 layout elements in the magazine subset of $M^6$ dataset}
\label{fig6}
\end{figure}

\subsection{A.2 Alignment of PDF to HTML}

The mapping of text from PDFs back to HTML is performed using the heuristic rules outlined in Algorithm 1, including text content cleaning, template construction for alignment, coarse matching based on textual similarity (e.g., edit distance) and spatial constraints, followed by fine-grained alignment leveraging contextual coherence, among others. The complete implementation details and source code are provided in the supplementary materials.

\begin{algorithm}[h]
\caption{PDF-to-HTML Text Alignment}
\label{alg:alignment_html}
\begin{algorithmic}[1]
\Require $\mathcal{L}$: PDF lines $\{(t_l, \mathbf{b}_l)\}$, 
         $\mathcal{C}$: HTML blocks $\{(c_j, \tau_j, \mathbf{p}_j)\}$
\Ensure $\mathcal{D}$: Aligned structure $[j][\tau][p] \mapsto$ lines

\State $\texttt{pool} \gets \texttt{BuildPool}(\mathcal{C})$ \State \Comment{Clean \& index crawled texts}
\State $\mathcal{D} \gets \text{dict}(|\mathcal{C}|+1)$ \State \Comment{Init; last bin for unmatched}

\For{each $l \in \mathcal{L}$ from block $b$}
    \State $t_l^c, t_b^c \gets \texttt{PrepareTexts}(l, b)$ \State \Comment{Clean line/block texts, drop prefixes}
    \State $\texttt{cands} \gets \texttt{FindCandidates}(t_l^c, t_b^c, l.\mathbf{b}, \texttt{pool})$ \State \Comment{Filter by text, spatial, and context similarity}
    
    \State $\text{match} \gets \texttt{None}$
    \If{$|\texttt{cands}| = 1$}
        \State $\text{match} \gets \texttt{cands}[0]$
    \ElsIf{$|\texttt{cands}| > 1$}
        \State $p, n \gets \texttt{GetNeighborTexts}(b.id)$ \State \Comment{Get prev/next PDF block texts}
        \State $s(c) \gets \max(\texttt{Sim}(p + c.t, t_b^c), \texttt{Sim}(c.t + n, t_b^c))$
        \State $c^* \gets \argmax_{c \in \texttt{cands}} s(c)$ \State \Comment{Find best candidate via context}
        \If{$s(c^*) > 0.3$}
            \State $\text{match} \gets c^*$
        \EndIf
    \EndIf

    \If{$\text{match} \neq \texttt{None}$}
        \State $(j, \tau) \gets (\text{match}.j, \text{match}.\tau)$
    \Else
        \State $(j, \tau) \gets (|\mathcal{C}|, \texttt{OtherText})$  \State\Comment{Assign to unmatched bin}
    \EndIf
    \State $\mathcal{D}[j][\tau][b.id].\texttt{append}(l)$
\EndFor

\State \Return $\mathcal{D}$
\end{algorithmic}
\end{algorithm}

\subsection{A.3 Order Modeling for Document}

In constructing MosaicDoc, we adopted a nuanced approach to reading order annotation, recognizing that visually rich documents rarely have a single, linear reading path. A single page in a newspaper or magazine often contains multiple independent articles or content blocks, which a reader might consume in various sequences. Therefore, drawing inspiration from precedents like the ROOR dataset, we refrained from enforcing a single, global reading order across an entire page. Instead, our methodology focuses on establishing robust local reading orders at the line and paragraph levels, wherever a definitive sequential or hierarchical relationship exists. To handle the diverse document types in our dataset, we developed three distinct, tailored strategies.

\noindent\textbf{Strategy for Chinese Newspapers}\hspace{1.3em}
we leverage the fact that many official publisher websites provide structured HTML versions of each page, complete with article-level metadata such as titles, subtitles, bylines, captions, and precise layout bounding boxes. This enables us to reconstruct accurate reading orders by aligning the text extracted from the PDF images with the corresponding HTML content. We employ the context-aware text alignment method described in Section A.2, which uses fuzzy matching based on edit distance, constrained by spatial proximity and contextual coherence, to map OCR-derived text spans to their canonical HTML positions. The sequence of text elements in the HTML source then serves as the ground truth for line- and paragraph-level reading order. It is important to note that the HTML structure and metadata formatting vary significantly across publishers. Therefore, we implement publisher-specific scraping pipelines and fine-tune the bounding box regression to accurately map HTML elements back to their corresponding regions in the original document images.

\noindent\textbf{Strategy for Bilingual Magazines}\hspace{1.3em}
As illustrated in Figure 7, a similar HTML-based strategy was applied to our collection of bilingual magazines. First, we mapped the text from the source HTML back to the corresponding layout elements on the document page using contextual edit distance matching. Subsequently, we established the reading order by inferring connections between these elements based on a combination of two signals. Semantic Sequence: The natural order of text as it appears in the HTML source code.Layout Hierarchy: A set of predefined precedence rules governing the relationships between different layout element types (e.g., a title must precede a paragraph; a caption must follow a figure). The directed graph in Figure 7 visualizes these hierarchical constraints, where an arrow indicates a valid sequential relationship. This hybrid approach, which combines the semantic sequence from HTML with visual-structural rules, allows us to reliably model the reading order even in complex, non-Manhattan layouts common in magazines.

\begin{figure}[t]
\centering
\includegraphics[width=0.45\textwidth]{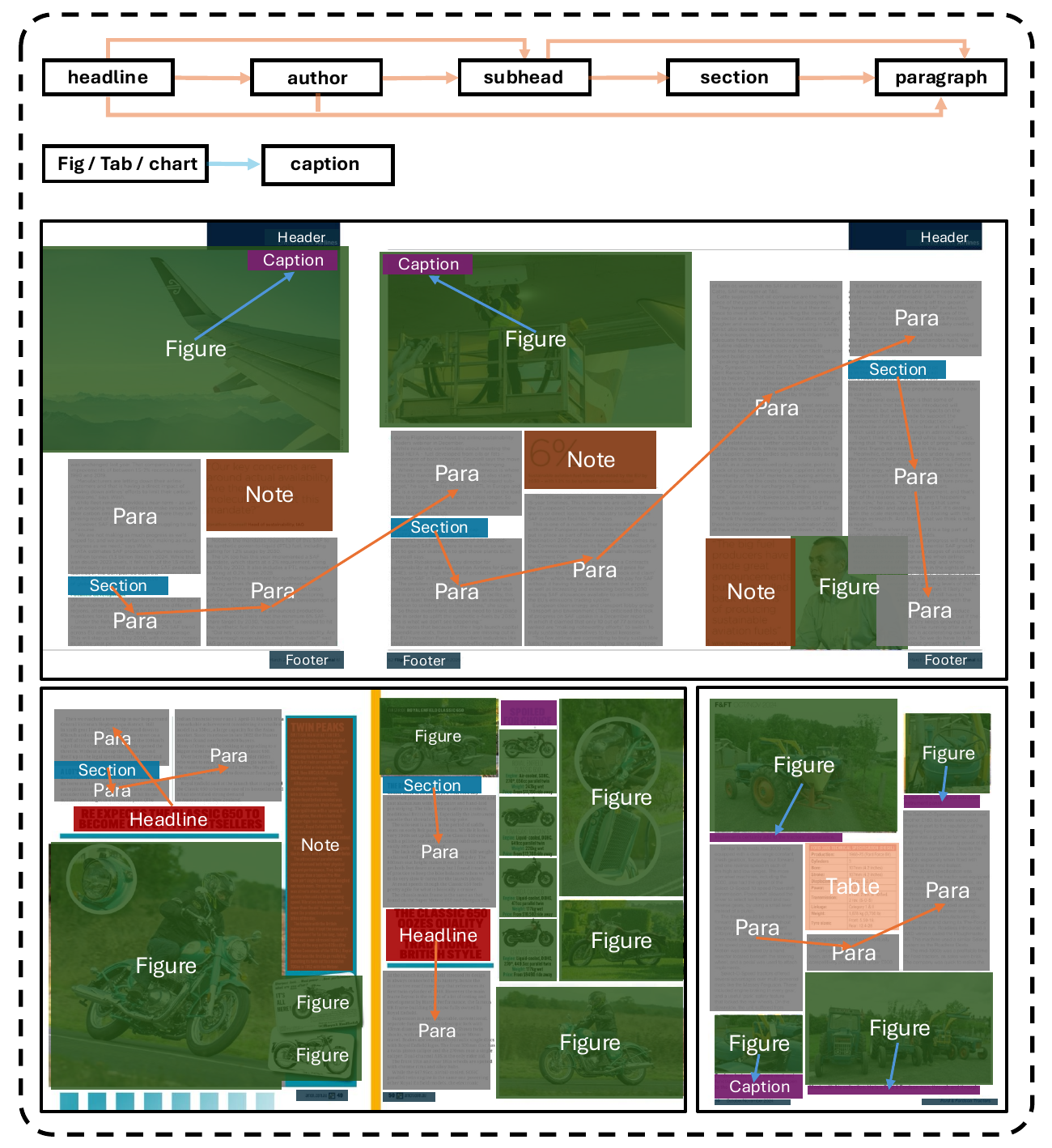}
\caption{Hierarchical reading order constraints among layout elements in bilingual magazines}
\label{fig7}
\end{figure}

\begin{figure}[t]
\centering
\includegraphics[width=0.45\textwidth]{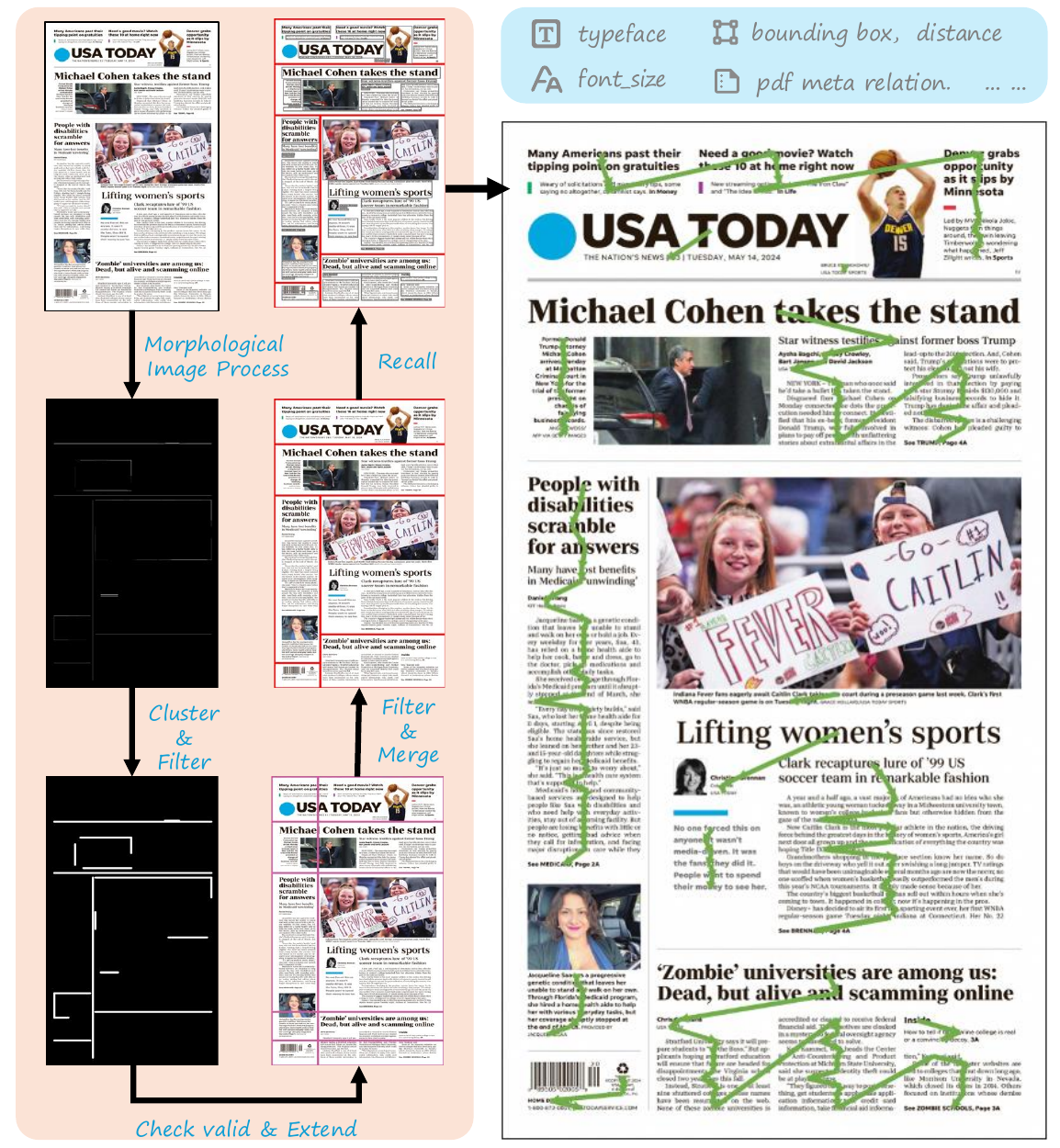}
\caption{Pipeline for reading order inference in English newspaper layouts}
\label{en_news_order}
\end{figure}

\noindent\textbf{Strategy for English Newspapers}\hspace{1.3em}
The aforementioned methods were not applicable to English newspapers in our dataset, as they typically lack publicly available, structured metadata. Furthermore, existing document parsing tools and rule-based methods perform poorly on their dense and varied layouts. Consequently, we developed a more robust, vision-centric pipeline to establish article-level reading order, as depicted in Figure~\ref{en_news_order}. We first apply a series of image processing techniques to segment the page into distinct article blocks. This involves binarization and morphological operations (dilation and erosion) to detect separating lines, followed by a rule-based filtering and merging process to identify the boundaries of each logical article. Once the articles are isolated, we determine the reading order within each block. This is achieved by leveraging a combination of PDF metadata cues, including the spatial coordinates of text lines (e.g., top-to-bottom, left-to-right), font attributes (style and size), and the intrinsic text sequence embedded within the PDF structure. This multi-cue approach ensures a robust reading order inference even in the absence of external structured data.

\begin{table*}[ht]
\centering
\small
\setlength{\tabcolsep}{6pt} 
\renewcommand{\arraystretch}{1.2} 
\begin{tabular}{ccccccccccc}
\toprule
\multirow{2}{*}{\textbf{Subset}} & \multirow{2}{*}{\textbf{Image}} & \multicolumn{4}{c}{\textbf{DocVQA}} & \multicolumn{2}{c}{\textbf{ROP}} & \multicolumn{2}{c}{\textbf{OCR}} & \textbf{Grounding} \\
\cmidrule(l{12pt}r{10pt}){3-6} \cmidrule(l{10pt}r{11pt}){7-8}  \cmidrule(l{9pt}r{13pt}){9-10}  \cmidrule{11-11}
                & & \textbf{S} & \textbf{M} & \textbf{T} & \textbf{C} & \textbf{Line} & \textbf{Para} & \textbf{Lines} & \textbf{Words} & \textbf{Ans} \& \textbf{Cont} \\
\midrule
en-Magzine     & 215 & 1,116  & 707 & 141 & 110 & 32,847 & 4,344  & 37,093 & 203,696   & 1,821 \\
zh-Magzine     & 218 & 763    & 432 & 142 & 138 & 24,837 & 2,644  & 27,289 & 318,035   & 1,195 \\
en-Newspaper   & 220 & 840    & 267 & 127 & 92  & 79,365 & 5,035  & 86,302 & 399,142   & 1,107 \\
zh-Newspaper   & 215 & 855  & 1,234 & 122 & 179 & 86,161 & 12,669 & 91,868 & 1,410,104 & 2,089 \\
\bottomrule
\end{tabular}
\caption{Distribution of the MosaicDoc test subset. The table includes annotations for document visual question answering (DocVQA), line- and paragraph-level reading order prediction (ROP), line- and word-level OCR, and spatial grounding of answers and context (Ans \& Cont).}
\label{test_subset}
\end{table*}

\section{B. Prompt templates}

In this section, we present detailed implementations of the generator, Discriminator and guardrails in DocWeaver, including the design of their respective prompt templates.

\subsection{B.1 Prompt for Generator}

We employ GPT-4o and DeepSeek-R1 as the primary generators to produce questions from given textual content, HTML tables, and charts. After experimentation with various template designs, we finalized the prompt templates shown in Figures~\ref{fig:sm_prompt} through~\ref{fig:char_prompt}. Each template consists of five key components: (1) fine-grained task instructions, (2) a summary of the input content, (3) the input data along with relevant layout information, (4) specified output format requirements, and (5) eight in-context examples—four selected randomly from a dataset-based example pool and four from a human-constructed example pool. For text inputs, the layout block category of the text is included; for tables, both the image and the corresponding HTML structure along with extracted text are provided; for charts, the image and any embedded textual elements are supplied to the generator. The table template is designed with reference to the implementation in TVG~\cite{liu2024tvg}.

\subsection{B.2 Prompt for Discriminator}

To enhance question difficulty, we employ QWQ-32B as an auxiliary discriminator to determine whether the generated content contains inferable or computable information. If such information is identified, the context is fed back to the generator to produce multi-hop question-answer pairs. The prompt template used by the discriminator is illustrated in Figure~\ref{fig:discriminator_prompt}. It guides the model to assess the presence of structured, reasoning-enabling data (e.g., numerical values, relational patterns) within the input and to flag instances suitable for complex reasoning.

\subsection{B.3 Prompt for Guardrails}

The hallucination guardrails are also implemented using QWQ-32B, due to its strong reasoning capabilities and support for local deployment, which enables efficient processing of large-scale datasets like MosaicDoc while reducing operational costs. The architecture of each hallucination guardrail is depicted in Figure~\ref{fig:guardrails_prompt}. The prompt for each guardrail includes a carefully designed set of evaluation criteria and step-by-step instructions. The model is required to first read and understand the evaluation criteria, then follow the prescribed steps to assess the generated question-answer pair against the original document and its context, assigning a score that reflects factual consistency. These scores are used to filter the generated question-answer pairs to ensure dataset quality. A pair is considered valid only if all guardrails assign a score higher than 3; otherwise, if any score is less than or equal to 3, the pair is flagged as hallucinated and discarded.

\section{C. Dataset Curation and Statistics}

This section details the data curation pipeline for the newspaper subset of MosaicDoc, including automated cleaning strategies and test set composition.

\subsection{C.1 Geometric Consistency Filtering}

To ensure high-quality reading order annotations in the newspaper subset, we apply a rule-based filtering strategy to remove erroneous or implausible line-to-line connections. Let $\mathbf{p}_i = (x_i, y_i)$ and $\mathbf{p}_j = (x_j, y_j)$ denote the center coordinates of two text-line bounding boxes. A directed link from box $i$ to box $j$ is considered \textit{invalid} if it is both long and nearly horizontal—patterns inconsistent with typical reading flow and layout geometry.

We define $\mathcal{I}_{ij}$ as a binary indicator of such invalid connections:
\begin{equation}
\setlength{\arraycolsep}{3pt}
\mathcal{I}_{ij} \triangleq \left\{
\begin{array}{l}
\sqrt{(x_j - x_i)^2 + (y_j - y_i)^2} > T_1, \\[1mm]
\arccos\left( \dfrac{|x_j - x_i|}{\sqrt{(x_j - x_i)^2 + (y_j - y_i)^2}} \right) < \theta_1,
\end{array}
\right.
\end{equation}
where both conditions must hold simultaneously. Here, $T_1$ and $\theta_1$ are predefined thresholds on length and orientation, respectively. In our implementation, we set $T_1 = w$ (The wide margin pixels of image) and $\theta_1 = 45^\circ$, based on empirical analysis of document layout structures. This criterion effectively filters out spurious connections that arise due to layout parsing errors or ambiguous proximity.

A document is discarded if it contains more than five such invalid links, along with its associated reading order annotations. This step ensures that only documents with geometrically plausible structure are retained in the final dataset.

\subsection{C.2 Test Set Composition}
The test set of MosaicDoc was carefully curated through manual inspection to ensure both annotation accuracy and layout diversity. All reading order annotations were verified by at least two annotators, with inter-annotator agreement measured by an average pairwise F1 score of 0.97. The final distribution of the test set is presented in Table~\ref{test_subset}.

\begin{table}[t]
\centering
\footnotesize
\setlength{\tabcolsep}{6pt} 
\renewcommand{\arraystretch}{1.2} 
\begin{tabular}{ll}
\toprule
\textbf{Dataset} & \textbf{Tasks} \\ 
\midrule
MP-DocVQA                   & DocVQA, Line-level OCR \\
TAT-DQA                     & DocVQA, word-level OCR \\
InfographicsVQA             & DocVQA, line- and word-level OCR \\
DuReader$_{vis}$            & DocVQA, word-level OCR \\
DUDE                        & DocVQA, line- and word-level OCR \\
                            & Answer Spotting \\
ROOR                        & para-level ROP, line- and word-level OCR \\
ReadingBank                 & text-level ROP, word-level OCR \\
olmOCR-mix-0225             & text-level ROP \\
\midrule
\multirow{3}{*}{\textbf{MosaicDoc}~(ours)}   & DocVQA, line- and word-level OCR, \\
                            & line- and para-level ROP,   \\
                            & Answer~/~Context~/~Text Spotting \\
\bottomrule
\end{tabular}
\caption{Supported tasks across document understanding benchmarks. including document visual question answering (DocVQA), reading order prediction (ROP) at both line and paragraph levels, multi-granularity OCR, and spatial grounding of answers, context, and text.}
\label{benchmark_task}
\end{table}

\section{D. More Properties of MosaicDoc}

This section provides additional characteristics of the MosaicDoc dataset, offering deeper insights into its data sources, layout complexity, and task coverage.

\subsection{D.1 Document Source}

MosaicDoc exhibits high source diversity, encompassing materials from 196 distinct publishers across 24 different domains. The data is primarily collected from the official websites of major Chinese newspapers, aggregator platforms such as VKontakte\textsuperscript{10}\footnotetext[10]{\url{https://vk.com/}} and FreeMagazines.top\textsuperscript{11}\footnotetext[11]{\url{https://freemagazines.top/}}, as well as few original documents compiled by $M^6$ dataset.
\begin{itemize}
\item \textbf{The Chinese newspaper} subset is sourced from the official websites of leading national newspapers in China, including \textit{People's Daily}, \textit{Economic Daily}, \textit{Procuratorial Daily}, \textit{China Education Daily}, \textit{China Culture Daily}, \textit{Xinhua Daily}, and \textit{Yangcheng Evening News}. 
\item \textbf{The English newspaper} subset mainly comprises publications from prominent English-language newspapers such as \textit{The Wall Street Journal}, \textit{USA Today}, \textit{The Washington Post}, and \textit{The Los Angeles Times}. 
\item \textbf{The Chinese magazine} subset includes issues from 52 well-known periodicals, such as \textit{Aerospace World}, \textit{China National Tourism}, \textit{NBA Special Edition}, and \textit{Global Screen}. 
\item \textbf{The English magazine} subset is drawn from major publications including \textit{New Scientist}, \textit{The New Yorker}, \textit{Harvard Business Review}, and \textit{The Wall Street Journal Magazine}, along with over 130 additional magazines covering diverse topics.
\end{itemize}

\subsection{D.2 BLEU Scores of Subset}
\begin{figure}[t]
\centering
\includegraphics[width=0.5\textwidth]{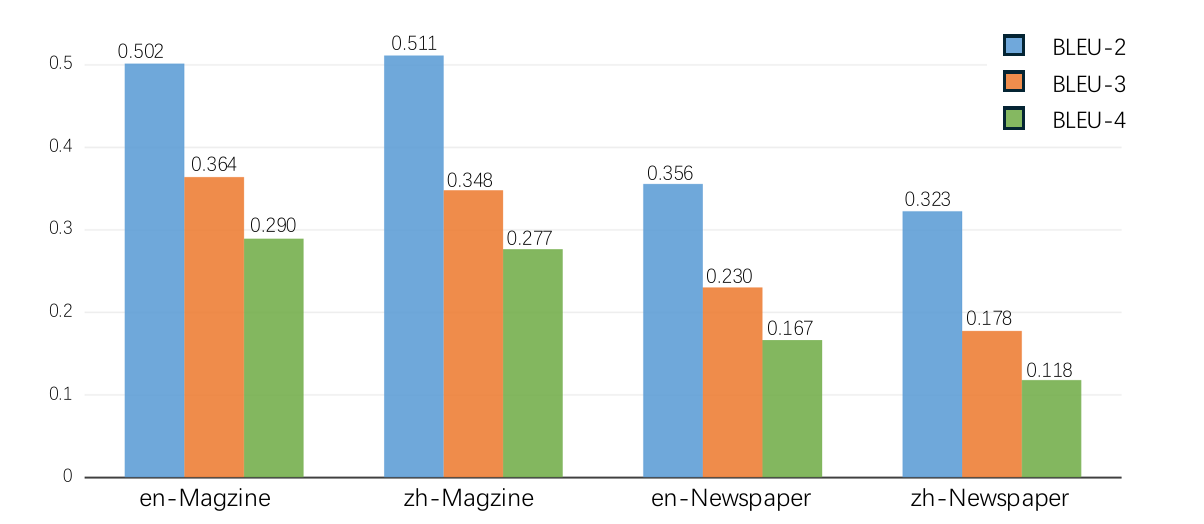}
\caption{The BLEU Scores on MosaicDoc Subsets.}
\label{bleu_subset}
\end{figure}

Figure~\ref{bleu_subset} presents the BLEU scores for each subset of MosaicDoc. To quantify layout complexity, we treat the globally correct reading order as the reference text, constructed by sorting all text lines in a document according to their top-left word coordinates in a left-to-right, top-to-bottom order. The model-generated reading sequence is similarly linearized as the candidate text. The BLEU score (up to 4-gram overlap) is then computed between the two, serving as a proxy for structural complexity—lower scores indicate more non-linear or intricate layouts.

Consistent with the evaluation in Figure~\ref{fig:bleu-bechmark}, we randomly sample 1,000 document images from each subset (or dataset), compute the BLEU score per document, and report the average. Note that BLEU-1 is always 1.0 due to the completeness of character coverage, and is thus omitted from visualization.

\subsection{D.3 Tasks of Each Benchmark}
Table~\ref{benchmark_task} summarizes the tasks supported by each benchmark included in the main experiments, serving as a complement to Table~\ref{tab: benchmarks} in the main text.

\section{E. Setting for Evaluation}

\subsection{E.1 LLM for Baselines}

Table~\ref{benchmark_llm} shows the types of LLMs used by each of the SOTA VLMs employed in the experiment.

\begin{figure}[t]
\centering
\includegraphics[width=0.5\textwidth]{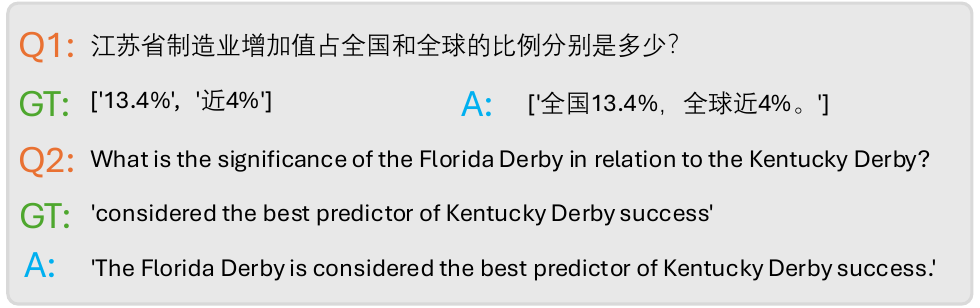}
\caption{Examples of questions and answers that should be regarded as positive examples.}
\label{fig:vqa_badcase}
\end{figure}

\begin{table}[t]
\centering
\footnotesize
\setlength{\tabcolsep}{3pt} 
\renewcommand{\arraystretch}{1.1} 
\begin{tabular}{ll|ll}
\toprule
\textbf{Model} & \textbf{LLM-base} & \textbf{Model} & \textbf{LLM-base} \\ 
\midrule
GOT-OCR                   & Qwen-0.5B & mPlug-DocOWL2                        & LLaMA-7B \\
olmOCR                     & Qwen2-7B & CogVLM2                        & Meta-Llama-3-8B \\
Vary                      & Qwen-7B & InternVL3                 & Internlm3-8B \\
TextMonkey              & Qwen-7B  & Qwen2.5-VL             & Qwen2.5-7B \\
\bottomrule
\end{tabular}
\caption{Corresponding LLM backbone of VLMs}
\label{benchmark_llm}
\end{table}

\subsection{E.2 Definition of Evaluation Metrics}
\noindent\textbf{Average Normalized Levenshtein Similarity for List~(ANLSL)}\hspace{1.3em}

ANLS was introduced in \cite{docvqa} for document VQA and extended to order-invariant list evaluation~\cite{tito2021doccvqa}. We adapt the underlying edit distance measure to avoid penalizing valid outputs from vision-language models (VLMs) that are character-wise correct but differ slightly in formatting—examples are shown in Figure~\ref{fig:vqa_badcase}. For simplicity, we first define the evaluation for a single non-list ground truth $G$ and prediction $\hat{P}$, with string lengths $|G|$ and $|\hat{P}|$, respectively.

\begin{equation}
\mathrm{LD}(G, \hat{P}) =
\begin{cases}
    |G| \quad \text{if } |\hat{P}| > 3 |G|, \\
    0 \quad \text{if } G \land \hat{P} = G, \\
    |G| \quad \text{if } |\hat{P}| = 0, \\
    \mathrm{LD}(\mathrm{tail}(G), \mathrm{tail}(\hat{P})) \quad \text{if } G[0] = \hat{P}[0], \\
    1 + \min
    \begin{cases}
        \mathrm{LD}(\mathrm{tail}(G), \hat{P}) \\
        \mathrm{LD}(G, \mathrm{tail}(\hat{P})), \text{ otherwise}\\
        \mathrm{LD}(\mathrm{tail}(G), \mathrm{tail}(\hat{P}))
    \end{cases}
\end{cases}
\label{ld}
\end{equation}

where $\mathrm{tail}(S)$ removes the first character. Then:
\begin{equation}
\mathrm{NLS}(G, \hat{P}) = 1 - \frac{\mathrm{LD}(G, \hat{P})}{\max(|G|, |\hat{P}|)},
\end{equation}

For lists $G = \{g_i\}_{i=1}^M$, $\hat{P} = \{p_j\}_{j=1}^N$, we compute pairwise $\mathrm{NLS}(g_i, p_j)$, solve optimal matching:
\begin{equation}
\mathcal{M}^* = \arg\min_{\mathcal{M}} \sum_{(i,j) \in \mathcal{M}} (1 - \mathrm{NLS}(g_i, p_j)),
\label{eq:assignment}
\end{equation}
and define:
\begin{equation}
\mathrm{ANLSL}(G, \hat{P}) = \frac{\sum_{(i,j) \in \mathcal{M}^*} \mathrm{NLS}(g_i, p_j)}{\max(M, N)} ,
\label{eq:anlsl}
\end{equation}
where $M = |G|$ and $N = |\hat{P}|$. This final score lies in $[0,1]$, with 1 indicating perfect alignment between all matched items.

\noindent\textbf{Output-based Character Recognition Rate~(OCRR)}\hspace{1.3em}
We use the Character Recognition Rate (CRR) to measure the proportion of correctly recognized characters in the model's output:
\begin{equation}
\mathrm{CRR} = \frac{1}{|G|} \sum_{(i,j) \in \mathcal{M}^*} \mathbf{1}[g_i = p_j],
\label{eq:crr}
\end{equation}

where $\mathcal{M}^*$ denotes the optimal matching set as defined in Equation~\ref{eq:assignment}, and $\mathbf{1}[\cdot]$ is the indicator function.
Considering that the model output frequently contains repetitive or skipped characters, we introduce the OCRR to further assess the output text quality:
\begin{equation}
\mathrm{OCRR} = \frac{1}{|\hat{P}|} \sum_{(i,j) \in \mathcal{M}^*} \mathbf{1}[g_i = p_j],
\label{eq:ocrr}
\end{equation}

\section{F. Qualitative Experiments}

This section presents qualitative results to visually illustrate model performance and limitations on key tasks.

\subsection{F.1 Document Visual Question Answering}

Figure~\ref{dovqa-visual} presents qualitative results for selected state-of-the-art (SOTA) models on the DocVQA task. 

\subsection{F.2 Spatial-aware Task}
We assess a model's ability to spatially ground VQA answers and their context by locating their coordinates. On the complex layouts in MosaicDoc, existing models fail at this task, yielding near-zero scores. To better probe this capability, we introduced two simplified tasks: text grounding (text to coordinates) and spatial-aware recognition (coordinates to text), using context lines adjacent to the answer as the target. Quantitative and qualitative results are shown in Table~\ref{tab:text_ground} and~\ref{tab:spotting_ocr} and Figure~\ref{fig:tg-go}.

\begin{table}[t]
\centering
\scriptsize
\setlength{\tabcolsep}{2.5pt} 
\renewcommand{\arraystretch}{1.1} 
\begin{tabular}{ccccccc}
\toprule
\multirow{2}{*}{\textbf{Model}}  & \multicolumn{3}{c}{\textbf{Magzine} (en / zh)} & \multicolumn{3}{c}{\textbf{Newspaper} (en / zh)}  \\
\cmidrule(l{5pt}r{6pt}){2-4} \cmidrule(l{5pt}r{6pt}){5-7}
                      & $\mathbf{P}$    & $\mathbf{R}$    & $\mathbf{F1}$    & $\mathbf{P}$ & $\mathbf{R}$  & $\mathbf{F1}$ \\ 
\hline
InternVL3         & 2.82/15.3    & 4.11/25.3    & 3.34/719.1  & 0.53/0.49    & 1.39/2.13  & 0.76/0.80 \\
Qwen2.5-VL       & 32.2/23.1    & 17.9/30.3    & 23.0/26.2  & 14.6/4.46    & 15.5/3.92  & 15.04/4.17 \\
\bottomrule
\end{tabular}
\caption{Text grounding results}
\label{tab:text_ground}
\end{table}

\begin{table}[t]
\centering
\scriptsize
\setlength{\tabcolsep}{6pt} 
\renewcommand{\arraystretch}{1.1} 
\begin{tabular}{ccccc}
\toprule
\multirow{2}{*}{\textbf{Model}}  & \multicolumn{2}{c}{\textbf{Magzine}} & \multicolumn{2}{c}{\textbf{Newspaper}}  \\
\cmidrule(l{5pt}r{6pt}){2-3} \cmidrule(l{5pt}r{6pt}){4-5}
                                & \textbf{EN}    & \textbf{ZH}    & \textbf{EN}    & \textbf{ZH}  \\ 
\hline
InternVL3         & 1.61    & 1.40    & 0.71  & 0.00  \\
Qwen2.5-VL       & 17.15    & 15.17    & 17.36  & 22.01   \\
\bottomrule
\end{tabular}
\caption{The ANLS score of spatial-aware recognition task}
\label{tab:spotting_ocr}
\end{table}

\subsection{F.3 Reading Order Prediction}

Figure~\ref{fig:rop-visual} shows the reading order prediction performance of SOTA models on complex layouts. A key finding from Tables 5 and 11 is the discrepancy between line-level and paragraph-level accuracy. Models excel at ordering individual lines but falter with paragraphs, especially those spanning multiple columns or that are visually adjacent but semantically unrelated.

\begin{table}[ht]
\centering
\scriptsize
\setlength{\tabcolsep}{2.5pt} 
\renewcommand{\arraystretch}{1.1} 
\begin{tabular}{ccccccc}
\toprule
\multirow{2}{*}{\textbf{Model}}  & \multicolumn{3}{c}{\textbf{Magzine} (en / zh)} & \multicolumn{3}{c}{\textbf{Newspaper} (en / zh)}  \\
\cmidrule(l{5pt}r{6pt}){2-4} \cmidrule(l{5pt}r{6pt}){5-7}
                      & $\mathbf{P}$    & $\mathbf{R}$    & $\mathbf{F1}$    & $\mathbf{P}$ & $\mathbf{R}$  & $\mathbf{F1}$ \\ 
\hline
olmOCR           & 87.3/89.1    & 11.7/10.9   & 20.6/19.4  & 64.3/19.8   & 4.96/7.29  & 9.21/10.7 \\
InternVL3         & 76.0/78.5    & 10.1/10.2    & 17.8/18.0  & 57.5/55.5    & 3.70/8.77  & 6.96/15.1 \\
Qwen2.5-VL       & 81.7/82.0    & 12.0/11.4    & 21.0/20.0  & 67.2/73.7    & 4.30/11.3  & 8.01/19.5 \\
GPT-4o       & 79.6/64.1    & 10.4/6.30    & 17.2/11.5  & 66.8/21.6   & 3.97/0.47  & 7.28/0.92 \\
Gemini-2.5       & 86.5/86.2    & 25.3/26.4    & 39.1/40.4  & 74.73/83.4   & 4.41/12.9  & 8.33/22.3 \\
\bottomrule
\end{tabular}
\caption{Paragraph level reading order prediction results}
\label{tab:para_rop}
\end{table}

\begin{figure*}[t]
\centering
\includegraphics[width=1\textwidth]{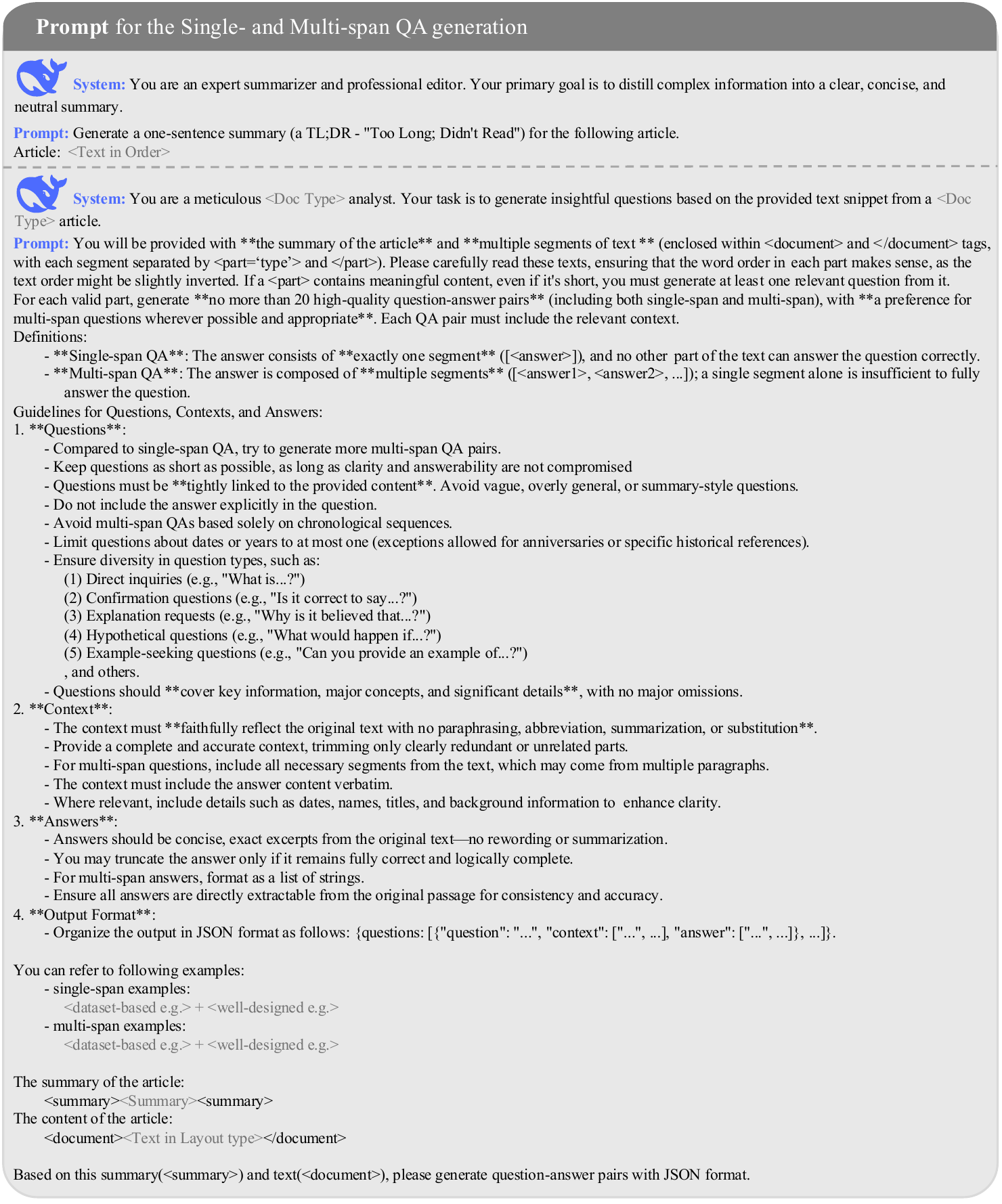}
\caption{The prompt for the Single- and Multi-span QA generation}
\label{fig:sm_prompt}
\end{figure*}

\begin{figure*}[t]
\centering
\includegraphics[width=1\textwidth]{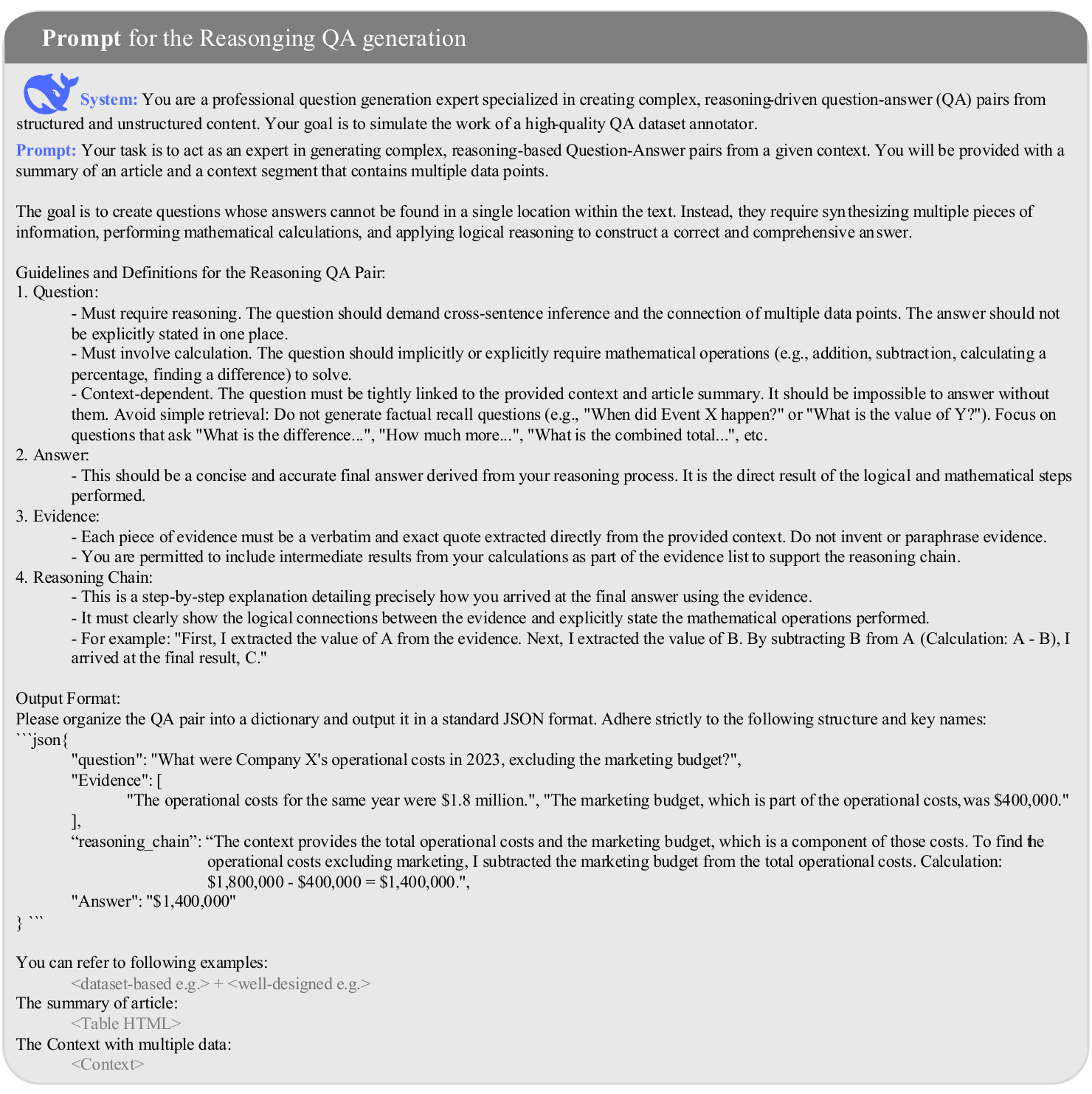}
\caption{The prompt for the Reasonging QA generation}
\label{fig:mh_prompt}
\end{figure*}

\begin{figure*}[t]
\centering
\includegraphics[width=1\textwidth]{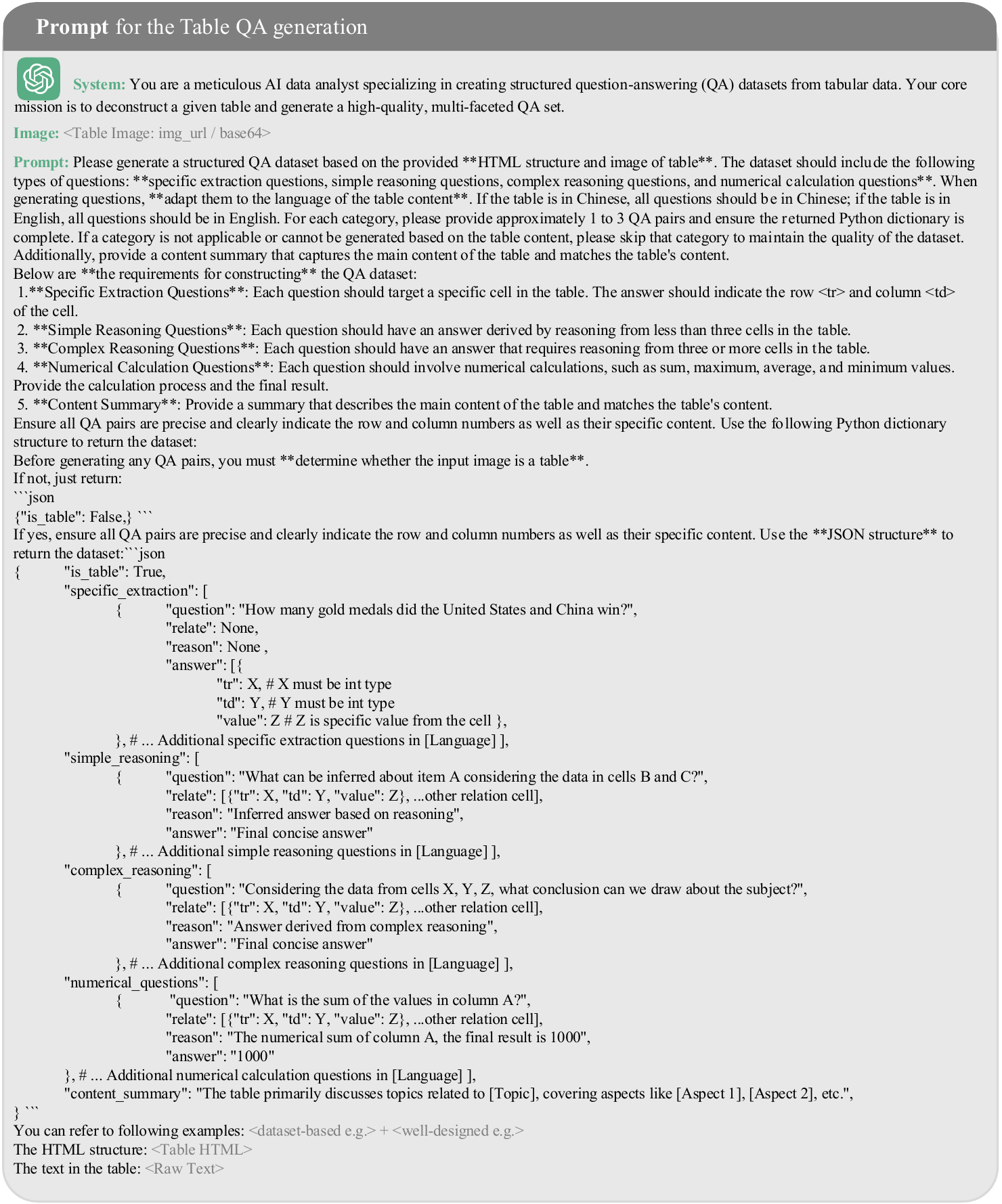}
\caption{The prompt for the Table QA generation}
\label{fig:tab_prompt}
\end{figure*}

\begin{figure*}[t]
\centering
\includegraphics[width=1\textwidth]{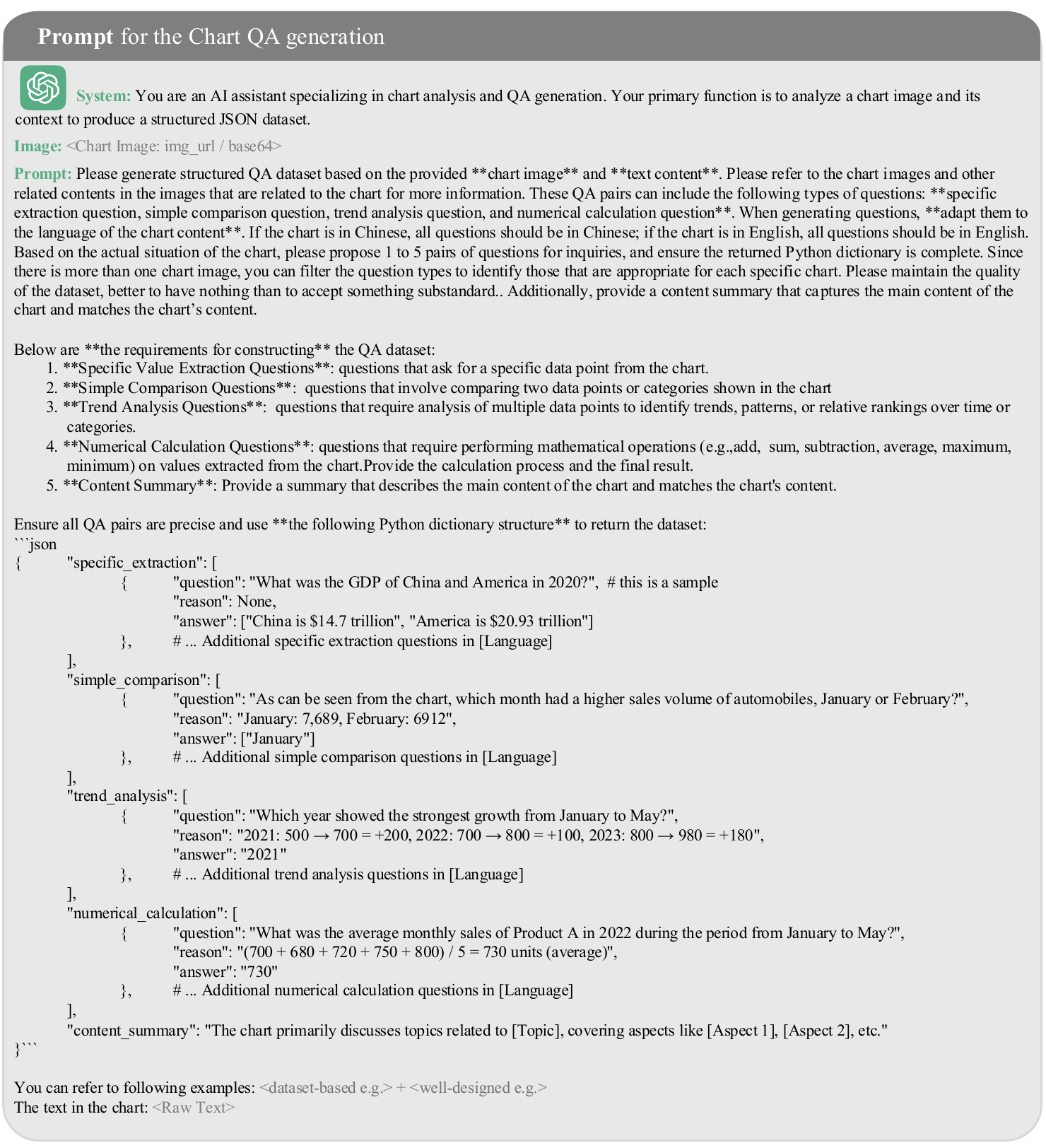}
\caption{The prompt for the Chart QA generation}
\label{fig:char_prompt}
\end{figure*}

\begin{figure*}[t]
\centering
\includegraphics[width=1\textwidth]{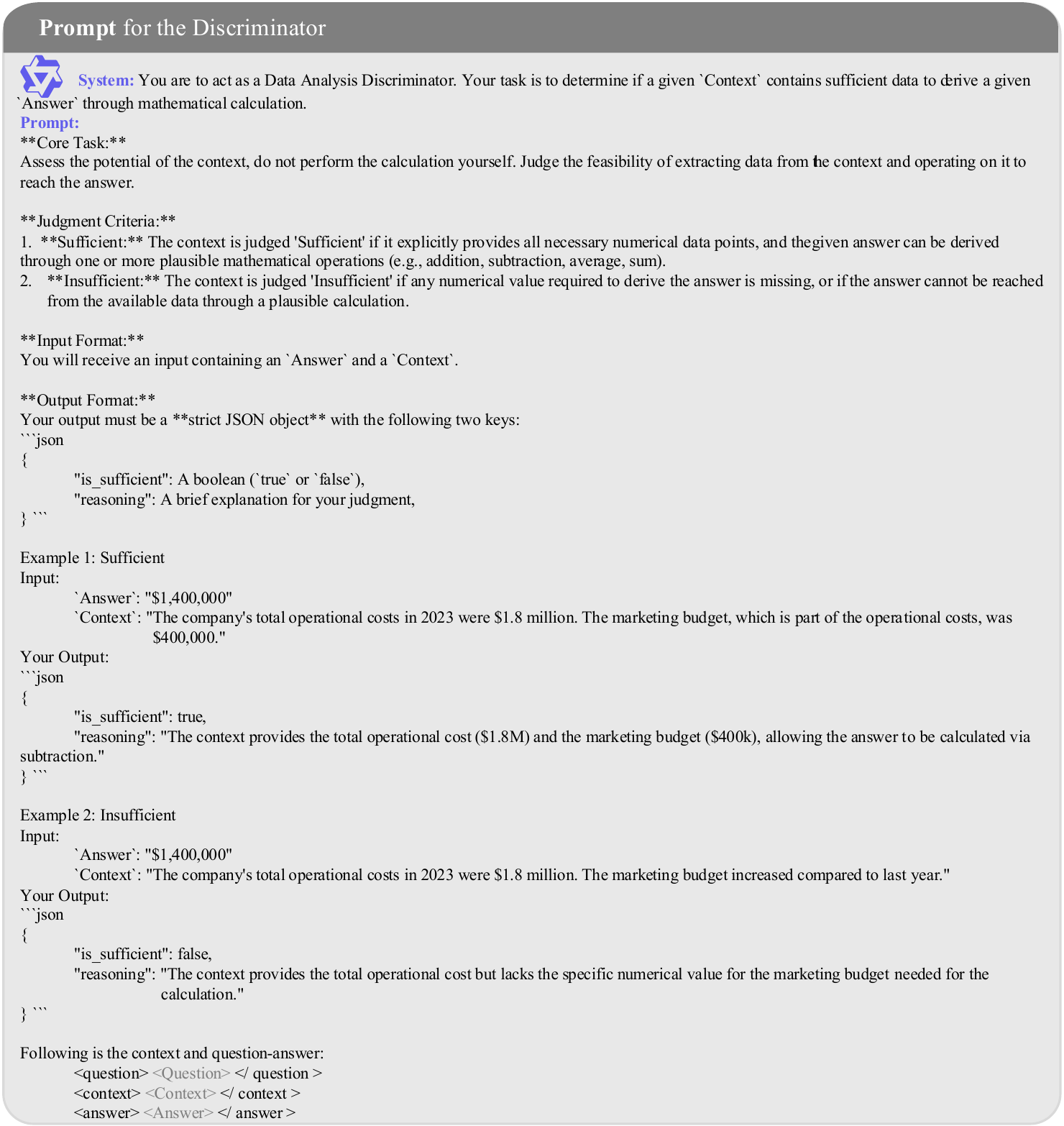}
\caption{The prompt for the Discriminator}
\label{fig:discriminator_prompt}
\end{figure*}

\begin{figure*}[t]
\centering
\includegraphics[width=1\textwidth]{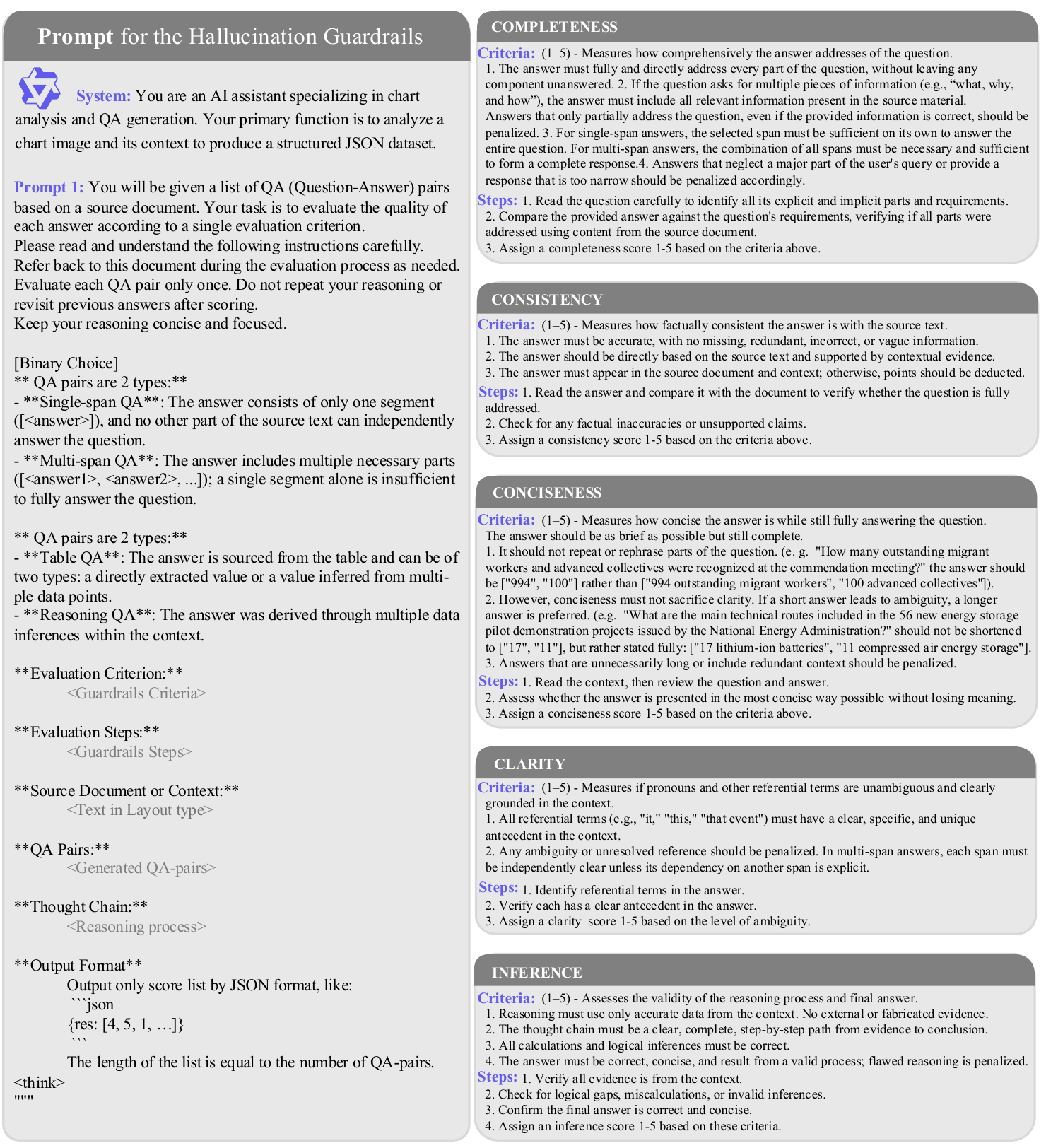}
\caption{The prompt for the Hallucination Guardrails with the criteria and steps}
\label{fig:guardrails_prompt}
\end{figure*}

\begin{figure*}[t]
\centering
\includegraphics[width=1\textwidth]{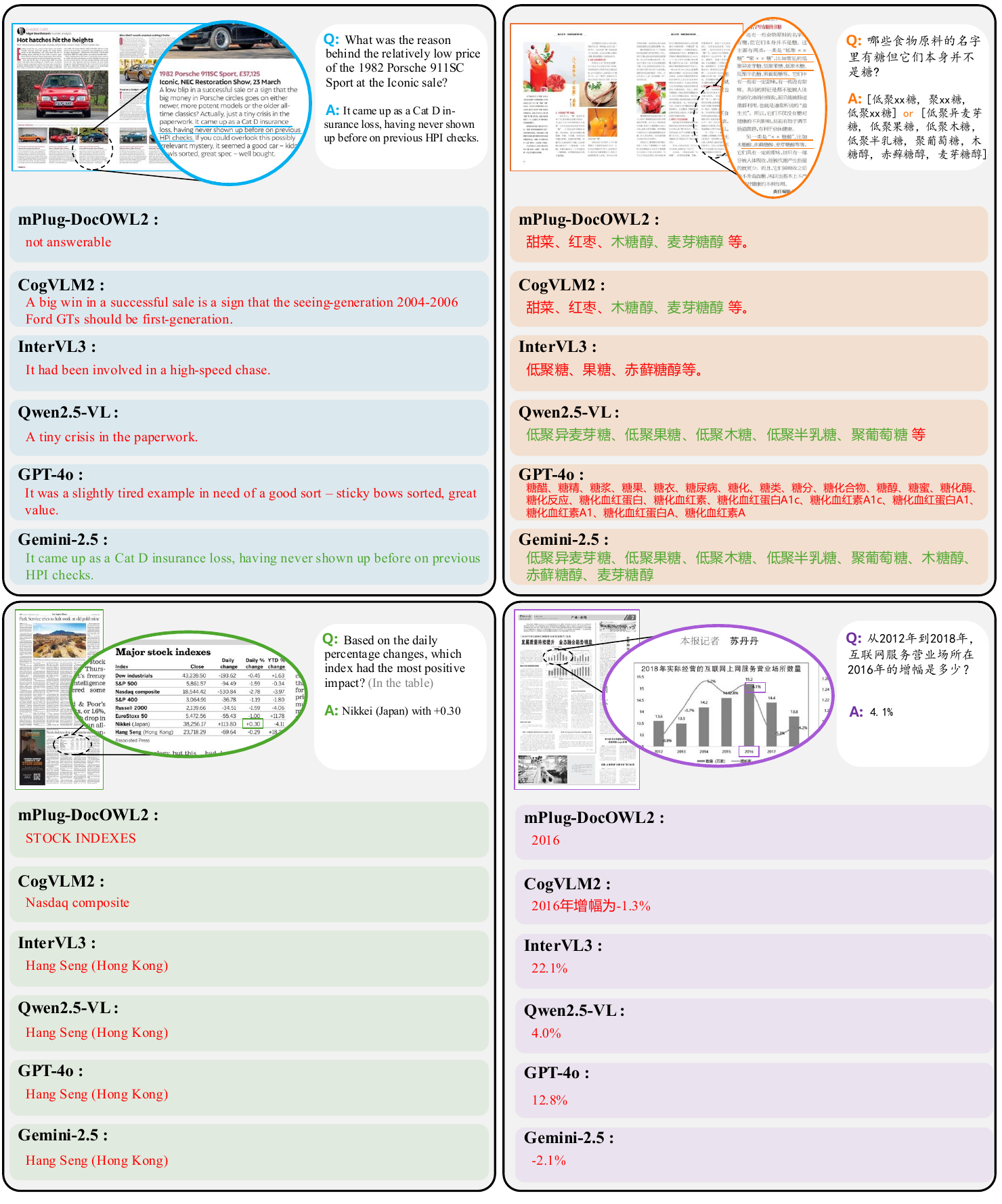}
\caption{Qualitative results of six state-of-the-art VLMs on the DocVQA task. The figure showcases four distinct question-answering scenarios (from top-left to bottom-right): single-span, multi-span, table-based, and chart-based}
\label{dovqa-visual}
\end{figure*}

\begin{figure*}[h] 
    \centering 
    \begin{subfigure}{1\textwidth} 
        \centering
        \includegraphics[width=\textwidth]{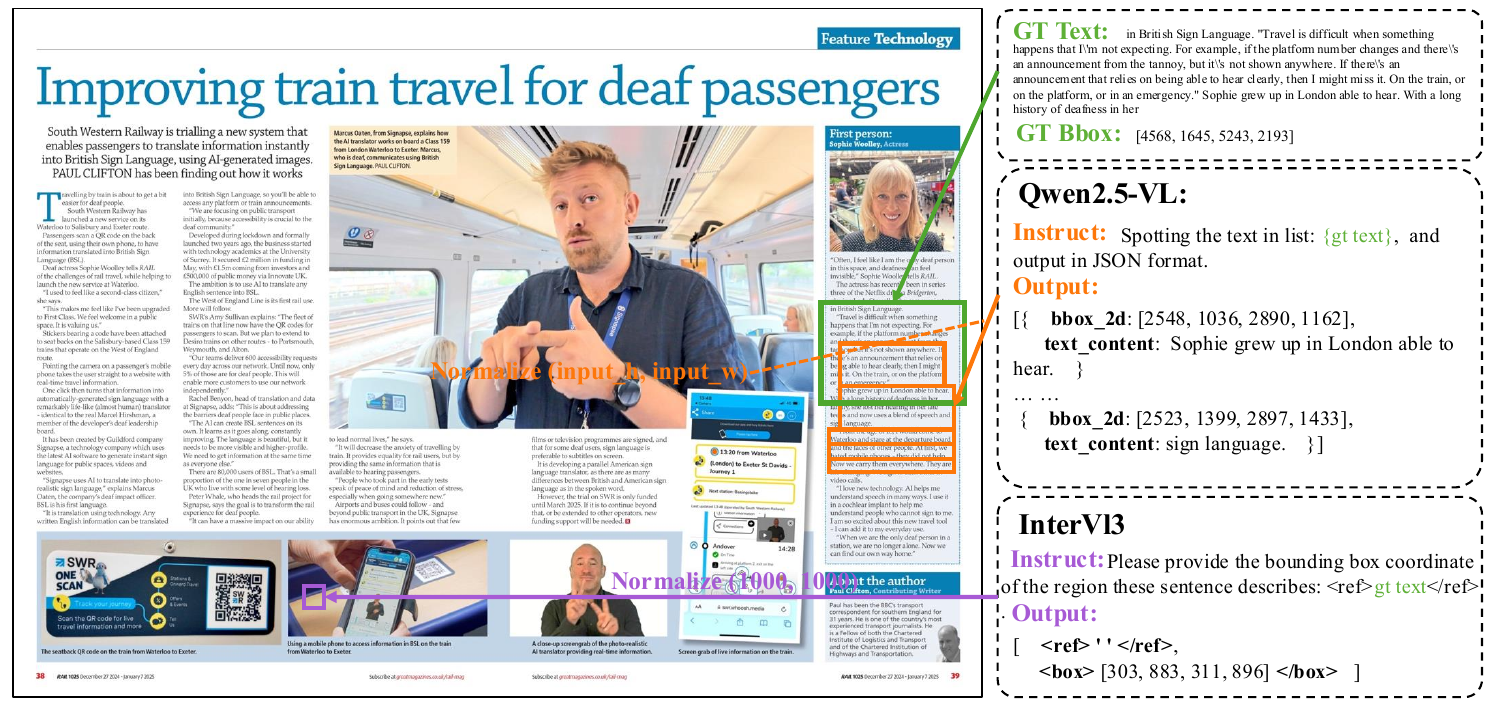} 
        \caption{Qualitative results on text grounding}
        \label{fig:tg-part}
    \end{subfigure}
    \hfill 
    \begin{subfigure}{1\textwidth}
        \centering
        \includegraphics[width=\textwidth]{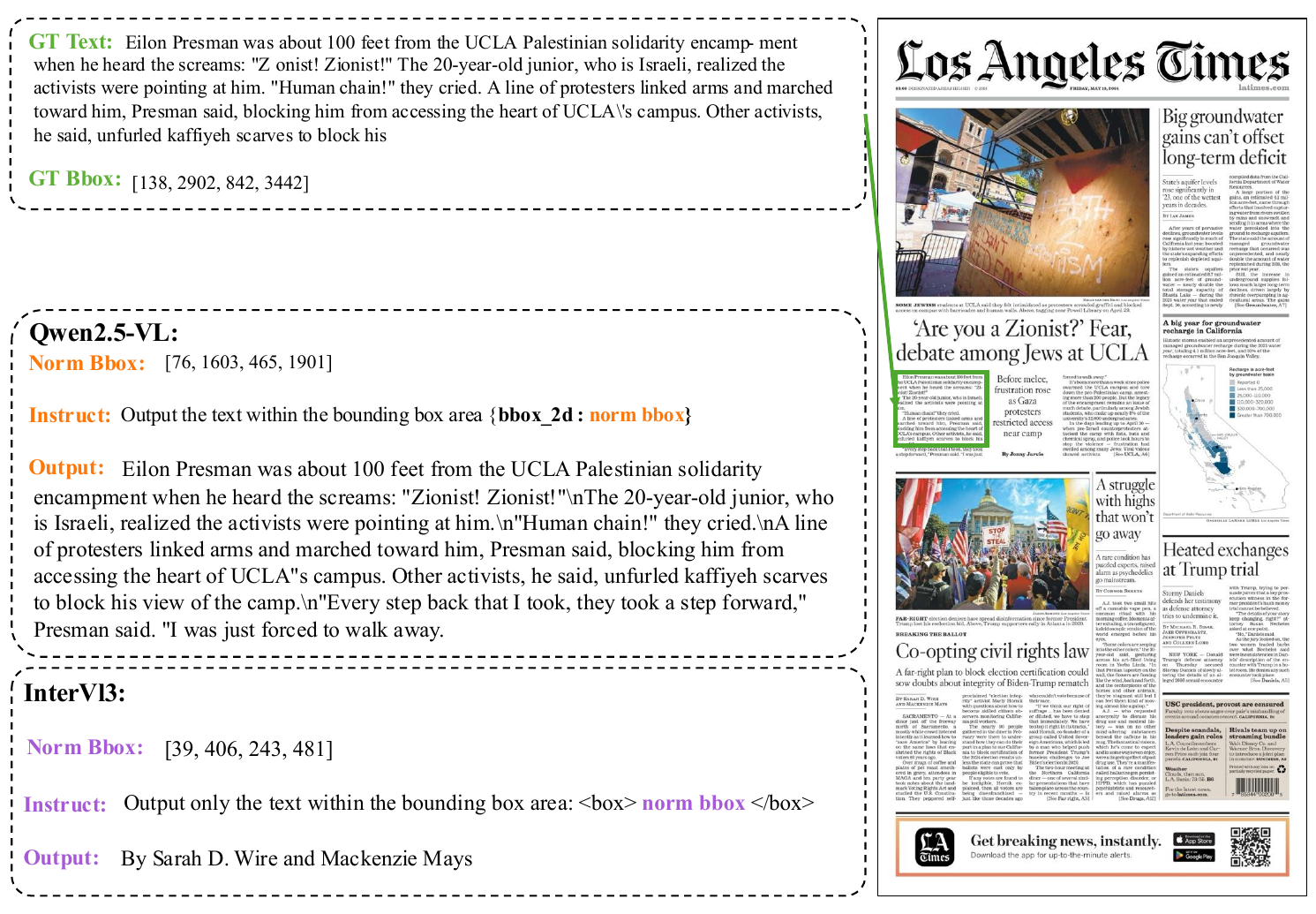} 
        \caption{Qualitative results on spatial-aware recognition}
        \label{fig:go-part}
    \end{subfigure}
    \caption{Qualitative results across Qwen2.5-VL and InternVL3 on spatial-aware tasks}
    \label{fig:tg-go}
\end{figure*}

\begin{figure*}[t]
\centering
\includegraphics[width=1\textwidth]{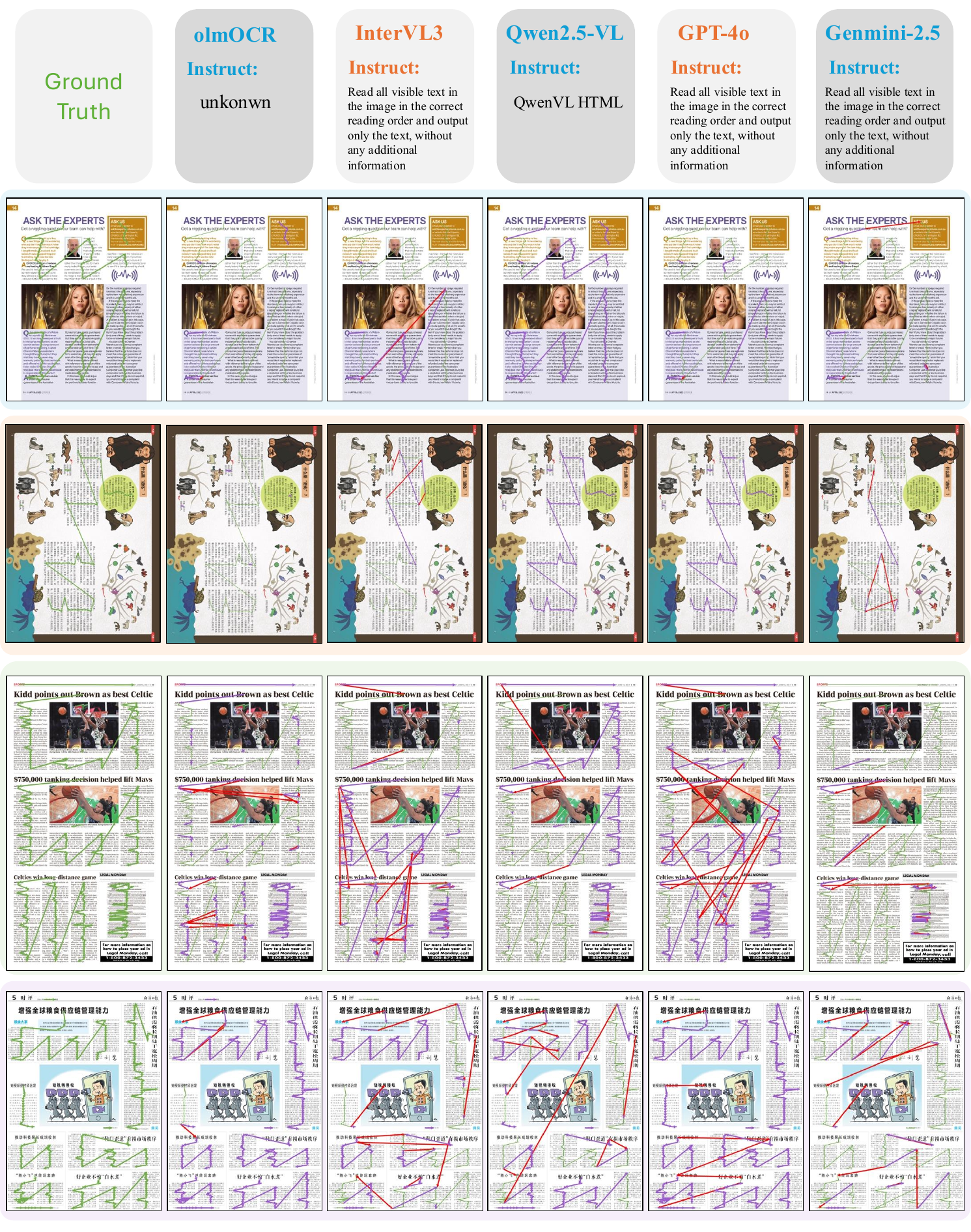}
\caption{Qualitative results for line-level reading order prediction, where colored arrows denote the outcome: \textcolor{green}{green} indicates correct predictions, \textcolor{red}{red} indicates incorrect ones, and \textcolor{purple}{purple} highlights missed relationships.}
\label{fig:rop-visual}
\end{figure*}

\end{document}